\documentclass{article}

\usepackage{microtype}
\usepackage{graphicx}
\usepackage{subcaption}
\usepackage{booktabs} 
\usepackage{hyperref}

\newcommand{\entry}[4]{%
  \node[font=\scriptsize] (mid) at (#1,#2) {$|$};
  \node[font=\scriptsize,left of=mid,anchor=east,node distance=0pt] {#3};
  \node[font=\scriptsize,right of=mid,anchor=west,node distance=0pt] {#4};}

\usepackage{icml2026}

\usepackage{amsmath}
\usepackage{amssymb}
\usepackage{mathtools}
\usepackage{amsthm}

\usepackage[dvipsnames]{xcolor}

\usepackage{tikz}
\usetikzlibrary{arrows.meta}

\usepackage[capitalize,noabbrev]{cleveref}

\theoremstyle{plain}

\theoremstyle{definition}

\theoremstyle{remark}

\usepackage[textsize=tiny]{todonotes}

\usepackage{wasysym}
\usepackage[utf8]{inputenc}
\DeclareUnicodeCharacter{03C3}{$\sigma$}
\DeclareUnicodeCharacter{2074}{$^4$}

\icmltitlerunning{LLMs with in-context learning for Algorithmic Theoretical Physics }

\begin{document}

\twocolumn[
\flushright\noindent\makebox{%
  \raisebox{0.5cm}[0pt][0pt]{\small IPMU26-0020}%
}%

\icmltitle{LLMs with in-context learning for Algorithmic Theoretical Physics} 

\center\textbf{ Anamaria Hell and Leander Thiele} 

  \begin{center}
      Kavli IPMU (WPI), UTIAS, The University of Tokyo,\\
and Center for Data-Driven Discovery,\\
5-1-5 Kashiwanoha, Kashiwa, Chiba 277-8583, Japan
  \end{center}

 \printAffiliationsAndNotice

  \vskip 0.3in
]

\begin{abstract}
  There is an increasing number of algorithmic computations in theoretical physics. These, while conceptually simple, can nevertheless be time-consuming and contain subtleties that should not be overlooked. Given the recent improvement of Large Language Models (LLM), it is natural to investigate whether LLMs equipped with a computer algebra system (CAS) runtime and sufficiently informative context can reliably carry out these algorithmic tasks. In this work, we interface Claude with Maple, and apply this framework to cosmological perturbations in modified theories of gravity. We demonstrate the current capabilities of this approach, the typical failures, and how the same can be improved.   
  We find that a frontier LLM supplied with worked examples is able to solve most test problems.
\end{abstract}

\section{Introduction}

Theoretical physicists deal with a range of tasks in their research.
Between the most mechanistic calculations which have been relegated to numerical and symbolic computer programs, and the art of constructing new theories born out of genuine creativity, lies a large field of ``algorithmic computations''.
By this we mean tasks that are sufficiently complex that a general solution in terms of a deterministic computer program is unavailable,
while at the same time not being of such difficulty that entirely new methods and ways of thinking are required for their solution.
To bracket algorithmic computations by examples, we mean tasks more complex than individual commands in a computer algebra system, but less complex than the discovery of AdS-CFT correspondence or certain proofs in Einstein gravity.

This leaves a large terrain of computations for which a generic recipe exists while the details vary from problem to problem and certain subtleties may require special care.
Perturbative calculations (like scattering matrices) in renormalizable QFT, string theory, or EFTs are typical examples.
More familiar to general physicists would be typical problems in classical electrodynamics or perturbation theory in quantum mechanics.

Such algorithmic computations currently take up much time for many theoretical physicists, even with the help of computer algebra systems (CAS).
At the same time, their relative simplicity implies that their solution using agentic AI might be within reach.
The algorithmic nature and frequency of these problems implies that a large set of worked examples is available and an abstract description of the algorithm can be written down.
In this work, we use both worked examples and the algorithm's description for in-context learning and test which contents in the context optimize success rate.

We thus test in this work the ability of a frontier large language model, interfaced with a CAS session and equipped with an informative context, to solve problems from a specific class of algorithmic computations:
\textit{identifying the degrees of freedom for cosmological backgrounds in various theories of gravity}.
Cosmological perturbations, small ripples in spacetime and matter, give rise to the structure of the Universe that we observe today \cite{Starobinsky:1979ty, Mukhanov:1981xt,  Chibisov:1982nx, Kodama:1984ziu, Sasaki:1986hm, Mukhanov:1990me}. They represent the degrees of freedom, the building blocks of theories, which determine their stability and observational consequences. Identifying them is thus essential, and can be non-trivial when departing from Einstein's General Relativity.  Several approaches have been developed with this purpose, including the Hamiltonian analysis with a Dirac-Bergmann algorithm \cite{Bergmann:1949zz, Dirac:1950pj, Anderson:1951ta}, and a \textit{Faddeev-Jackiw} approach \cite{Faddeev:1988qp}, as well as Lagrangian formalism (see \citet{Hell:2026blj} for a general prescription, and references therein), and a more rigorously mathematical approach founded directly in the equations of motion (see \citet{Heisenberg:2025fxc} and references therein).  In this work, we follow the \textit{recipe} given in \citet{Hell:2026blj}, which allows for the easiest and most straight-forward implementation in the context of cosmological perturbations.

For the purposes of a rigorous study, algorithmic computations of cosmological perturbations have some distinct advantages.
It is easy to construct test problems with research-grade complexity and with such obscure features that they are impossible to have been encountered in the LLMs training set.
In fact, in our evaluation we sometimes observe the LLM attempting to relate the task to some better-known problem from the literature.
At the same time, the type of problems have unique and easily checked solutions. 

Our work aims to establish how much theoretical physicists can benefit from automated computation specifically in algorithmic problems.
To this end, we answer the following questions:
\begin{enumerate}
  \item \textit{What are current capabilities of frontier LLMs equipped with a CAS runtime;}
  \item \textit{How should the LLM's context be used;}
  \item \textit{What are typical failure modes and how can they be mitigated.}
\end{enumerate}
Thanks to the restricted set of problems and our available expert knowledge, we can understand the LLM's solution strategies beyond simply checking the final answer, thus enabling a complementary qualitative evaluation.
A central aim of this work is to develop practical guidelines for how practitioners can employ their existing worked examples as useful context for LLMs to autonomously tackle similar problems.

As we demonstrate in Sec.~\ref{sec:related}, most work on LLMs equipped with a symbolic computation tool has concentrated on mathematics.
Compared to these works, we focus less on the model's creativity in finding proof strategies and more on its ability to perform relatively routine computations that are simply burdensome for the human physicist.
Note, however, that these computations are not necessarily easy, and the use of a CAS does not guarantee correctness.
Another unusual aspect of this work is the difficulty in defining a locally relevant context: some manipulations need to be carried out with later steps in mind (for example, solution of background equations may produce roots if done naively, which would lead to problems in subsequent steps).
Thus, algorithmic theoretical physics is an interesting testing ground for long contexts.

The remainder of this paper is structured as follows.
In Sec.~\ref{sec:related}, we discuss previous works applying LLMs equipped with CAS to mathematics and physics problems.
In Sec.~\ref{sec:methods}, we describe our setup, experiments, and evaluation strategy.
In Sec.~\ref{sec:results}, we summarize our results both qualitatively and quantitatively.
We conclude in Sec.~\ref{sec:concl}.

\section{Related Work}\label{sec:related}

Using structure in an LLM's response goes back to chain-of-thought~\cite{wei2023chainofthoughtpromptingelicitsreasoning}.
A natural extension of this idea is to interface the model with tools~\citep[e.g.,][]{gao2023palprogramaidedlanguagemodels,chen2023programthoughtspromptingdisentangling}.
Such tool-aided reasoning has found widespread application in mathematics, thanks to the strict verifiability afforded by programs such as Lean~\citep[e.g.,][]{gou2024toratoolintegratedreasoningagent,khaitan2025oforgellmcomputer,breen2025axproverdeepreasoningagentic,nezhad2026symcodeneurosymbolicapproachmathematical,luo2026agentmathempoweringmathematicalreasoning}.
Not surprisingly, tool use is found to be especially helpful for more difficult problmes~\cite{das2024mathsenseitoolaugmentedlargelanguage}.
In a similar fashion to this work, in-context learning~\citep[e.g.,][]{dong2024surveyincontextlearning} has been studied as a way to improve the LLM's performance in math problems~\citep[e.g.,][]{liu2024makesincontextlearningeffective,sultan2026neurosymbolicapproachreliableproof,gao2026processincontextlearningenhancing}.
In contrast to explicitly fine-tuned models, general purpose models with useful context are easier to use in a human-AI collaboration and can be more easily adapted if features in the tools (e.g., Lean strategies) change~\cite{breen2025axproverdeepreasoningagentic}.

In the context of theoretical physics, using LLMs for automated problem solving has so far been explored less than in mathematics.
Part of the reason may be the difficulty in finding research-grade but still easily gradeable problems; recently there have been efforts towards constructing such benchmarks~\cite{chung2025theoreticalphysicsbenchmarktpbench,pan2026cmtbenchmarkbenchmarkcondensedmatter} and using them to evaluate test-time scaling~\cite{gao2025testtimescalingtechniquestheoretical}.
These works find that without dedicated tools the current frontier models are unable to solve most research-grade benchmark problems.
A recent paper that is similar in spirit to our work is \citet{menzo2026agenticdiagrammaticaautonomoussymbolic}: they let an LLM use a specialized Mathematica package to compute Feynman diagrams.
In contrast to our work, their LLM operates in a much more constrained environment: its only real decision is about a small number of binary choices to do with conventions.
Our work is focused on problems that require more actual intelligence to come up with calculation strategies and deal with subtleties.
Much more ambitiously, some recent works~\citep[e.g.,][]{agrawal2026fermiaccagentsparticletheory} attempt to let artificial agents perform the entire pipeline of theoretical physics research.
Others argue that this approach is unlikely to be successful~\cite{lu2026theoreticalphysicsresearchbenefit}.
We do not aim to replace the human researcher entirely and thus restrict the experiments to algorithmic theoretical physics problems.
Similarly to \citet{lu2026theoreticalphysicsresearchbenefit}, we anticipate that equipping the LLM with a CAS is likely useful.

One of the main aims of this work is to establish how well a current frontier model can use examples or other information provided in the context.
Indeed, it has been noted that long contexts can degrade performance~\citep[e.g.,][]{li2024longcontextllmsstrugglelong,liu2023lostmiddlelanguagemodels,jin2024longcontextllmsmeetrag,du2025contextlengthhurtsllm,huang2024fewermoreboostingllm}, even specifically for mathematics~\cite{liu2024makesincontextlearningeffective}.
These findings suggest that a small set of well-chosen example problems~\citep[for how to choose, e.g.,][]{yang2024learninganalogyenhancingfewshot} should optimize performance.
On the other hand, \citet{zou2025manyshotincontextlearninglongcontext} find a mixed picture and \citet{agarwal2024manyshotincontextlearning} note that more capable modern frontier models are able to better deal with larger contexts.

\section{Methods}\label{sec:methods}

In this section, we give a pedagogical introduction to the considered theoretical physics problems (Sec.~\ref{sec:dof}),
describe our implementation of an LLM equipped with a CAS runtime (Sec.~\ref{sec:repl}),
list the test problems (Sec.~\ref{sec:problems}),
describe our quantitative evaluation method (Sec.~\ref{sec:eval}),
and finally describe the different choices of context we provide to the LLM (Sec.~\ref{sec:ctx}).

\subsection{The basics of dof}\label{sec:dof}

In general, the number of degrees of freedom (dof) of a physical system is defined by the number of initial conditions to completely determine the Cauchy problem of equations of motion divided by two. For example, a free particle in 1+1 dimensions is described by the Lagrangian $L=\frac{1}{2}\dot{q}^2$ where dot is the derivative with respect to time, and $q$ is the generalized coordinate. By varying the corresponding action $S=\int dt L$ with respect to the generalized coordinate, we find the corresponding equation of motion $\Ddot{q}=0$. To completely determine the Cauchy problem, we need to determine two initial conditions -- the initial position and velocities $q(t_0)$ and $\dot{q}(t_0)$ -- which thus gives one dof. In quantum field theories and theories of gravity one more often studies the dynamics of fields. In contrast to a free particle, such objects also depend also on the point in space, with a prime example being the scalar field $\phi(\Vec{x},t)$. Formally speaking, a theory of a standard free canonical scalar field in four dimensions thus has an infinite number of dof, and so when we state this, we will refer to their number per point in space-time. 

In flat and cosmological backgrounds, the space-time has spherical symmetry which gives rise to three well-known types of perturbations that we classify according to the group of spatial rotations: scalar, vector and tensor modes. 
However, if we consider a theory of gravity, where the main quantity in addition to matter is the metric tensor, $g_{\mu\nu}$, not all of its components will be physical degrees of freedom. For example, general relativity has two dof instead of ten which correspond to all components of the metric tensor, meaning that the remaining ones are not propagating. While there are many ways to show that this holds, in this work, we focus only on actions that are quadratic in the perturbations around particular backgrounds based on the Lagrangian formulation, and determine the physical modes according to the procedure described in \cite{Hell:2026blj}.

\subsection{Implementation}\label{sec:repl}

Our experimental setup is relatively simple: the context and problem formulation are input via the LLM's API, which produces CAS commands to be locally executed in the runtime read-eval-print loop (REPL), CAS results are appended to the message history and sent back via the API, and so forth until the LLM decides a solution has been found (or aborts).
We choose to work with a single artificial agent since the considered computations are relatively linear and we do not expect a multi-agent environment would yield substantial improvements.
We call the API directly instead of relying on a ready-made SDK (e.g., Claude Code) since reproducible model ability is important for our experiments.
We choose to work with a fixed, typically rather long, context for multiple reasons.
First, it enables us to systematically study the impact of different materials in the context.
Second, frontier models are getting increasingly better with long contexts and have reached impressive capabilities already~\cite{agarwal2024manyshotincontextlearning}.
Third, given the specialized nature of the algorithmic computations we are aiming at, we anticipate practitioners to want a simple prescription how to apply automated computation for their own work which would usually rely on private example calculations.
Fourth, the considered computations can despite their algorithmic linear nature benefit from a ``look-ahead'', as discussed in the introduction.

For the LLM, we use Claude Opus 4-6, and for the CAS we use Maple.
It is important to turn the human-oriented printing off on Maple (\texttt{prettyprint=0}).
We find that the LLM relatively frequently restarts to a fresh Maple session when it notoices that it is stuck; we include this for our quantitative analysis below.
Most of our results are obtained with \texttt{max\_tokens=1024} and thinking turned off; we discuss experiments with thinking turned on in Sec.~\ref{sec:thinking}.
We run each problem only once because we observe relatively stable behavior when repeating some runs.
To control runtime, we set a maximum limit of 100 turns at which point the REPL gets aborted.
Empirically, we find that the turns limit interrupts the loop in situations that do not seem to be leading to success.

\subsection{Problems and restrictions}\label{sec:problems}

To analyse the usefulness of Claude in studying the cosmological perturbations, we evaluate it on a set of nine problems, given by the following:

\textbf{R2Fs} Consider $\frac{M_P^2}{2}R+\beta R^2$ gravity on a cosmological background,
where $M_P$ is the Planck mass, and $\beta$ is the coupling. Find the quadratic
action for the scalar perturbation around this background, and express it in
terms of the propagating modes.
Perform the calculation in the Jordan frame, you are not allowed to use the Einstein frame.

\textbf{sRFs} Consider $\frac{M_P^2}{2}(R-2\Lambda)+\frac{1}{2}\sigma^2(m^2+f(R))$ gravity on a cosmological background,
where $M_P$ is the Planck mass, $\Lambda$ is the cosmological constant, $\sigma$ is the scalar field, $m^2$ is the mass, and $f(R)$ is a function of Ricci scalar. Find the quadratic action for the scalar perturbations around this background, and express it in terms of the propagating modes.

\textbf{sRFv} Consider $\frac{M_P^2}{2}(R-2\Lambda)+\frac{1}{2}\sigma^2(m^2+f(R))$ gravity on a cosmological background,
where $M_P$ is the Planck mass, $\Lambda$ is the cosmological constant, $\sigma$ is the scalar field, $m^2$ is the mass, and $f(R)$ is a function of Ricci scalar. Find the quadratic action for the vector perturbations around this background, and express it in terms of the propagating modes. 

\textbf{sRFt} Consider $\frac{M_P^2}{2}(R-2\Lambda)+\frac{1}{2}\sigma^2(m^2+f(R))$ gravity on a cosmological background,
where $M_P$ is the Planck mass, $\Lambda$ is the cosmological constant, $\sigma$ is the scalar field, $m^2$ is the mass, and $f(R)$ is a function of Ricci scalar. Find the quadratic action for the tensor perturbations around this background, and express it in terms of the propagating modes. 

\textbf{sRMs} Consider $\frac{M_P^2}{2}(R-2\Lambda)+\frac{1}{2}\sigma^2(m^2+f(R))$ gravity on a flat background,
where $M_P$ is the Planck mass, $\sigma$ is the scalar field, $m^2$ is the mass, and $f(R)$ is a function of Ricci scalar. Find the quadratic action for the scalar perturbations around this background, and express it in terms of the propagating modes. 

\textbf{sRMt} Consider $\frac{M_P^2}{2}R+\frac{1}{2}\sigma^2(m^2+f(R))$ gravity on a flat background,
where $M_P$ is the Planck mass, $\sigma$ is the scalar field, $m^2$ is the mass, and $f(R)$ is a function of Ricci scalar. Find the quadratic action for the tensor perturbations around this background, and express it in terms of the propagating modes.  

\textbf{sRi2Ms} Consider $\frac{M_P^2}{2}(R-2\Lambda)+\frac{1}{2}\sigma^2(m^2+\xi R_{\mu\nu}R^{\mu\nu})$ gravity on a flat background, where $M_P$ is the Planck mass, $\sigma$ is the scalar field, $m^2$ is the mass, and $\xi$ is the coupling constant. Find the quadratic action for the scalar perturbations around this background, and express it in terms of the propagating modes.  

\textbf{sRi2Fs} Consider $\frac{M_P^2}{2}(R-2\Lambda)+\frac{1}{2}\sigma^2(m^2+\xi R_{\mu\nu}R^{\mu\nu})$ gravity on a cosmological background, where $M_P$ is the Planck mass, $\sigma$ is the scalar field, $m^2$ is the mass, and $\xi$ is the coupling constant. Find the quadratic action for the scalar perturbations around this background, and express it in terms of the propagating modes. 

\textbf{sRi2Ft} Consider $\frac{M_P^2}{2}(R-2\Lambda)+\frac{1}{2}\sigma^2(m^2+\xi R_{\mu\nu}R^{\mu\nu})$ gravity on a cosmological background, where $M_P$ is the Planck mass, $\sigma$ is the scalar field, $m^2$ is the mass, and $\xi$ is the coupling constant. Find the quadratic action for the tensor perturbations around this background, and express it in terms of the propagating modes. 

We note that the above problems are passed in plain text, and only set in an appropriate \LaTeX\ form for readability.
An actual input looks like the following example: 

{\small\texttt{\detokenize{Consider Mp^2/2*R+beta*R^2 gravity on a cosmological background, where Mp is the Planck mass, and beta is the coupling. Find the quadratic action for the scalar perturbation around this background, and express it in terms of the propagating modes. Perform the calculation in the Jordan frame, you are not allowed to use the Einstein frame.
}}}

In creating these problems, we only focus on tensor, or scalar-tensor theories in flat and cosmological backgrounds. A more comprehensive study including also other type of theories, such as vector, tensor or two-form theories will be presented in follow-up work. Aside from the first problem (R2Fs), remaining problems are built on the recently introduced constrained scalar-field framework\footnote{See also \cite{Hell:2026sxt} for a theory with constrained vector field.} \cite{Hell:2025lgn}, ensuring that their analysis was not previously worked out in the literature. The problems also test Claude on a wide range of issues, commonly found in studying the perturbations, such as finding special background solutions that give rise to multiple branches, resolving the constraints, dealing with higher-derivatives and reducing their order, as well as analyzing different types of the perturbations. R2Fs, in contrast, corresponds to the Starobinsky inflation, which is one of the most well-known models in Cosmology, and serves as a reference point.

\subsection{Output evaluation}\label{sec:eval}
We base our evaluation of the results on a \textit{pass/fail} basis. The solutions to the problems can be divided into four main steps: the setup, computing the background equations of motion, finding the second-order action, and analyzing the perturbations. Theories that also involve higher-derivative terms have an additional step in reduction of the derivatives that can happen either in the setup or at the level of perturbations. 

Based on this, we adopt the following rules for the evaluation:
\begin{enumerate}
  \item If the setup is wrong, it is a fail.
  \item If the background equations of motion are not computed, or not properly substituted in the action, and this affects the result, then it is a fail. Otherwise, if this does not change the result, it is a pass.
  \item The perturbations, and the reduction of the higher-order derivatives should be analyzed correctly for a pass.
  \item If the resulting code obtains a correct final result, but forgets a subtlety, then it is still a pass.
\end{enumerate}
In  order to illustrate the output, we present it for one of the problems in the Appendix~\ref{app:example}.

\subsection{Contexts}\label{sec:ctx}
In the spirit of in-context learning, we supplement our Claude-Maple interface with four choices of context: 

\textbf{Context ``10ex": A set of 10 solved examples, 60k tokens}
As context, we prepare 10 step-by-step solved and explained examples that would represent problems of finding propagating dofs of theories of gravity in cosmology and flat space-time. For illustration, an explicit presentation of one of these examples is given in the Appendix~\ref{app:context}. Most of these examples are just standard examples, meant to cover different simplifications and Maple-code when studying scalar, vector or tensor perturbations. Specially, they do not involve any higher derivatives but are at most quadratic in the fields. 

\textbf{Context ``3broad": A set of 3 solved examples, 18k tokens}
We test the case when the number of examples is overall reduced to three. This also reduces the overall cost of the project. To still keep it general enough, we select three representatives, corresponding to the Example 2, Example 3 and Example 8, which is stated in Appendix~\ref{app:context}.

\textbf{Context ``3tailored": A modified set of 3 solved examples, 24k tokens}
As we will see in the results, attempts to some problems with contexts 10ex and 3broad fail with relatively common mistakes. To examine this further, we keep the Example 3 and Example 8 as part of the input data, and supplement the two with another modified example that is supposed to account for the common mistakes, given in Sec.~\ref{TSc}.

\textbf{Context ``instruction": A general description of the method, 2k tokens}
Finally, we provide as an input just a general instruction text based on the recent work \cite{Hell:2026blj}, and with no example code. This \textit{recipe} is given in the Appendix \ref{app:instruction}.

\section{Results}\label{sec:results}

\begin{figure}
\centering
\begin{tikzpicture}[
    x=1.1cm,
    y=0.55cm,
    font=\small,
    guide/.style={gray!60, dashed, line width=0.5pt},
    fail/.style={circle, draw=Cerulean, fill=white, line width=0.8pt, minimum size=2.2mm, inner sep=0pt},
    pass/.style={circle, draw=Cerulean, fill=Cerulean, minimum size=2.2mm, inner sep=0pt}
]

\foreach \y/\label in {1/R2Fs,2/sRFs,3/sRFv,4/sRFt,5/sRMs,6/sRMt,7/sRi2Ms,8/sRi2Fs,9/sRi2Ft} {
    \draw[guide] (1,\y) -- (4,\y);
    \node[left] at (0.8,\y) {\label};
}

\foreach \x/\label in {1/10ex, 2/3broad, 3/3tailored, 4/instruction} {
    \draw[guide] (\x,1) -- (\x,9);
    \node[left, rotate=90] at (\x,0.6) {\label};
}

\foreach \x in {1,2,3,4} {
    \foreach \y in {1,...,9} {
        \node[fail] at (\x,\y) {};
    }
}

\node[pass] at (1,1) {};
\node[pass] at (2,1) {};
\node[pass] at (3,1) {};

\node[pass] at (2,2) {};

\node[pass] at (1,3) {};
\node[pass] at (2,3) {};
\node[pass] at (3,3) {};

\node[pass] at (1,4) {};
\node[pass] at (2,4) {};
\node[pass] at (3,4) {};
\node[pass] at (4,4) {};

\node[pass] at (3,5) {};

\node[pass] at (1,6) {};
\node[pass] at (3,6) {};
\node[pass] at (4,6) {};

\node[pass] at (1,7) {};
\node[pass] at (3,7) {};
\node[pass] at (4,7) {};

\node[pass] at (2,8) {};

\node[pass] at (3,9) {};

\node[pass] at (4.8,8.4) {};
\node[right] at (4.95,8.4) {pass};

\node[fail] at (4.8,7.7) {};
\node[right] at (4.95,7.7) {fail};
\end{tikzpicture}

\caption{Results for the problems depending on the provided context. }
\label{fig:pass-fail-grid}
\end{figure}
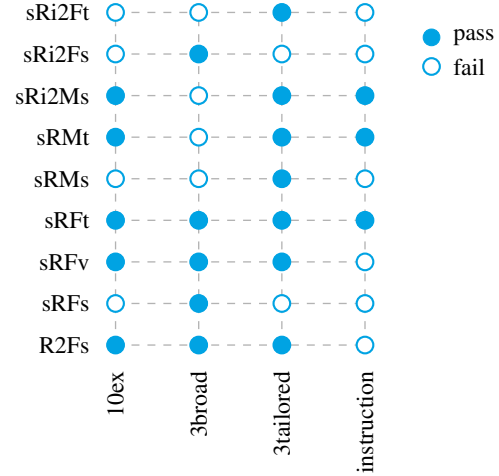

\begin{figure}
\centering

\begin{tikzpicture}[
    x=1.6cm,
    y=0.7cm,
    font=\small,
    guide/.style={gray!60, dashed, line width=0.5pt}
]

\foreach \y/\label in {1/R2Fs,2/sRFs,3/sRFv,4/sRFt,5/sRMs,6/sRMt,7/sRi2Ms,8/sRi2Fs,9/sRi2Ft} {
  
    \node[left] at (0.6,\y) {\label};
}

\foreach \x/\label in {1/10ex, 2/3broad, 3/3tailored, 4/instruction} {
    \node[left, rotate=90] at (\x,0.6) {\label};
}

\entry{1}{1}{49}{2}
\entry{1}{2}{82}{2}
\entry{1}{3}{32}{2}
\entry{1}{4}{57}{1}
\entry{1}{5}{93}{7}
\entry{1}{6}{18}{1}
\entry{1}{7}{37}{3}
\entry{1}{8}{66}{2}
\entry{1}{9}{69}{1}

\entry{2}{1}{75}{2}
\entry{2}{2}{89}{2}
\entry{2}{3}{32}{1}
\entry{2}{4}{37}{2}
\entry{2}{5}{\textcolor{Red}{100}}{6}
\entry{2}{6}{36}{1}
\entry{2}{7}{40}{2}
\entry{2}{8}{38}{1}
\entry{2}{9}{66}{2}

\entry{3}{1}{49}{1}
\entry{3}{2}{\textcolor{Red}{100}}{4}
\entry{3}{3}{30}{1}
\entry{3}{4}{29}{1}
\entry{3}{5}{55}{2}
\entry{3}{6}{19}{1}
\entry{3}{7}{24}{1}
\entry{3}{8}{77}{2}
\entry{3}{9}{41}{1}

\entry{4}{1}{50}{4}
\entry{4}{2}{95}{13}
\entry{4}{3}{79}{9}
\entry{4}{4}{73}{17}
\entry{4}{5}{70}{8}
\entry{4}{6}{28}{7}
\entry{4}{7}{28}{4}
\entry{4}{8}{80}{9}
\entry{4}{9}{\textcolor{Red}{100}}{7}

\begin{scope}[shift={(0,10)},y=0.03cm]
\draw[step={(1,20)},very thin,color=gray,dotted] (0.5,0) grid (4.5,90); 
\draw[->,Cerulean](0.5,0) -- (0.5,95) node[above,Cerulean] {turns (\newmoon)};
\foreach \y in {0,20,40,60,80} {
    \draw [Cerulean](0.55,\y) -- (0.45,\y) node[left,Cerulean] {\y};
}
\draw[Cerulean] plot[mark=*, mark size=2] coordinates {(1,55.88888889) (2,57.) (3,47.11111111) (4,67.)};
\draw[dashed,Cerulean] plot[mark=*, mark size=2] coordinates {(1,57) (2,40) (3,41) (4,73)};

\end{scope}

\begin{scope}[shift={(0,10)},y=0.3cm]
\draw[->,Peach](4.5,0) -- (4.5,9.5) node[above,Peach] {restarts ($\times$)};
\foreach \y in {0,2,4,6,8} {
    \draw[Peach] (4.45,\y) -- (4.55,\y) node[right,Peach] {\y};
}
\draw[Peach] plot[mark=x, mark size=3] coordinates {(1,2.33333333) (2,2.11111111) (3,1.55555556) (4,8.66666667)};
\draw[Peach,dashed] plot[mark=x, mark size=3] coordinates {(1,2) (2,2) (3,1) (4,8)};

\end{scope}

\end{tikzpicture}

\caption{The lower panel shows the solution length as turns$|$restarts depending on context (horizontal) and problem (vertical).
The upper panel summarizes the table in terms of mean (solid) and median (dashed).}
\label{fig:turns-restarts-grid}

\end{figure}
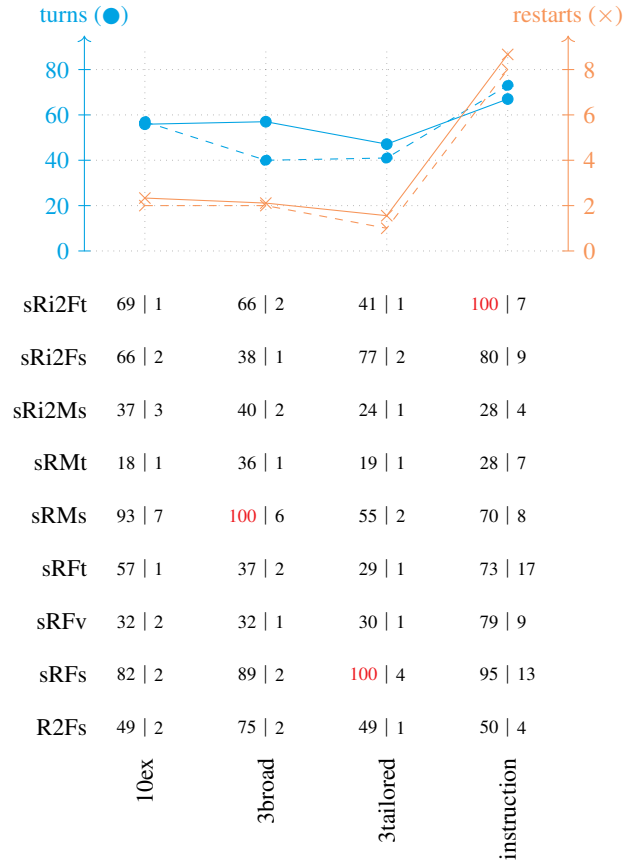

In Fig.~\ref{fig:pass-fail-grid}, we visualize the solutions' evaluation according to problem (vertical axis) and context (horizontal axis).
As an additional quantitative evaluation, we also provide the number of turns and number of Maple restarts in Fig.~\ref{fig:turns-restarts-grid}.
The baseline case of a long context (10ex) shows 5 correct solutions and 4 incorrect ones.
For the shorter context of representative problems (3broad), the success rate is identical but the failures mostly shift (there is a subtlety in this case to be discussed below).
The case of specifically tailored problems (3tailored), which is informed by the previous two experiments, shows the highest success rate and is the only one where sRi2Ft is solved correctly.
The 3tailored experiments also display the lowest number of turns and restarts on average, indicating that the deliberately constructed examples help the model find the solution efficiently.
The experiments with an abstract description of the \textit{recipe} (instruction) show the lowest number of successes and the highest average number of turns and restarts.
For a given problem, a higher number of turns and restarts is correlated with lower success rate.

In the following, we discuss the overview following each of the provided contexts, described in the previous section, followed by a brief comparison of thinking mode for selected problems.

\subsection{Context 10ex}

Overall, the performance of Claude with the submitted context has a marginal pass, with five problems correctly solved among nine. The number of turns -- sequences in which Claude runs Maple -- ranges from the minimal 18 to 93, with an average of 56, while the number of restarts\footnote{The minimal number of restarts should be one, corresponding to the start of the code.} is mostly from 1 to 3, with a special exception of 7 restarts in sRMs, and averaging on 2.

Among the given problems, R2Fs is a very useful reference in confirming that Claude uses not only the examples submitted in the context, but also information fed during training. In particular, while its task is to study the Starobinsky model in the Jordan frame, it still introduces an external scalar field, rewriting the initial action instead as $\int \sqrt{-g}\left(\frac{M_P^2}{2}+2\beta\sigma\right)R-\beta\sigma^2$ which reduces to the original theory by resolving the constraint for $\sigma$, and substituting it back to the action. This significantly simplifies the analysis, as it removes all of the higher derivatives acting on the perturbations that would otherwise be present until the order is reduced.

Throughout the remaining problems, one of the major challenges is the presence of higher-order derivatives. They were mostly dealt with through an incorrect substitution, with the aim of reducing the order of the higher derivatives acting on the perturbations. In particular, the main issue was that Claude forgets to include the constraints via the Lagrange multipliers, and thus reduces the order incorrectly. This was the case for sRFs, sRi2Fs and sRi2Ft.

Another challenge is resolving the background equations of motion. The above test problems were constructed with the intention of having multiple branching solutions, and thus challenging the LLM in this way as well. Surprisingly, however, different branches were noticed for all but sRMt, with Claude then selecting one of the branches (usually the more complicated one). The main issue associated to the background equations of motion was to resolve both constraint equation, acceleration equation and the equation for the scalar in such a way that there are no roots appearing in the solutions. In other words, the challenge is how to pick the variables which would be best to solve the equations for, when substituting them back to the action. Interestingly, in sRFv and sRFt, Claude also struggled with the background equations of motion, gave up and went to the second-order action. Only once there, it went back and studied the background equations, analyzing if they can simplify the resulting action at that point. 

An especially interesting case is sRMs. It is first correctly solved for the flat background. However, then, Claude deems the result too trivial, and instead turns to the analysis in curved space-time. This is interesting, because in sRMt where the flat background is formulated in the same way everything is solved correctly.

\subsection{Context 3broad}
In this case, the number of solved problems is again five, although the correctly solved problems themselves do not match with the previous context. The average number of turns is similar to 10ex. However, sRMs reaches the 100-turn limit thus automatically terminating. The average number of restarts is two, with sRMs being again an exception with 6 restarts overall. 

R2Fs is solved in a very similar way as in the previous case. Interestingly, Claude then uses the same trick to also solve sRFs, rewriting the action of the problem as
\begin{equation*}
    \begin{split}
        S&=\int d^4x\sqrt{-g}\left[\left(\frac{1}{2}\sigma^2f'(\chi)+\frac{M_P^2}{2}\right)R-M_P^2\Lambda\right.\\&\left.+\frac{1}{2}\sigma^2\left(m^2+f(\chi)-\chi f'(\chi)\right)\right].
    \end{split}
\end{equation*} 
Strictly speaking, this action is equivalent to the previous one only if $f''(\chi)\neq 0$. However, aside from this subtlety, the problem is now solved correctlyunlike with context 10ex. Another correctly solved problem is sRi2Fs. While it has noticed different branching solutions depending on if the constrained scalar vanishes or not, it chooses the simpler one with $\sigma(t)=0$. This procedure is equally valid as choosing the arbitrary branch, although it simplifies the analysis significantly.   In addition, we should note that sRFv and sRFt are mostly correctly solved. However, it should be noted that background equations of motion were overall unsolved/ignored.  Since this still didn't affect the final expression too much, we graded it with a passing grade, but one should keep in mind that this could have easily led to the wrong result in more complicated scenarios. 

sRMs, sRMt and sRi2Ms primarily fail because Claude understands flat background for cosmological evolution. However it would have been correct to select the Minkowski space-time. In a realistic human-AI interaction, this issue would have been caught. In future work, we may adjust the evaluation to account for such cases. Finally, in sRi2Ft it fails due to incorrect reduction of higher-order derivative terms.

\subsection{Context 3tailored}\label{TSc}

The primary issues in the previous two contexts are
\begin{itemize}
    \item reducing the higher-derivatives on the level of perturbations,
    \item solving background equations of motion,
    \item interpretation of the background on which perturbation theory should be performed. 
\end{itemize}
To counter these issues, in this context we replace the Example~2 with the following example on the pure $R^2$ gravity: 

\textbf{Modified example:} Consider $\beta R^2$ on a cosmological background, where $\beta$ is the coupling constant. Find the quadratic action for the scalar perturbation around this background in the Jordan frame, and express it in terms of the propagating modes.

This problem inherently contains the higher-derivatives in the Jordan frame, which are reduced on the level of perturbations to show that the theory describes one scalar mode on cosmological backgrounds (in addition to the two tensor ones), with the aim of countering the first issue found in the previous analyses. In addition, further explanations and tips on how to solve the background equations of motion in more general cases are added to the context. With aim of avoiding misinterpretations between flat background and cosmology, we also add: ``\textit{Note that if the problem said ``flat background”, you should assume Minkowski space-time, and set $a(t)=1$ too.}"
Finally, in the previous sets, we had noticed that the LLM spent many turns manipulating the form of the final action. To reduce the number of turns, we add an additional instruction: \textit{``If you have reached the final action, do not simplify it further."}

Overall, the performance of this set is significantly improved compared to the previous cases. Out of nine problems, only two were not solved: sRFs has reached 100 turns, which is why it had stopped. The LLM also did not perturb the constrained scalar in addition to the metric perturbations. The latter issue was also present in the sRi2Fs, which was otherwise very close to the correct solution. In contrast to the previous contexts this context does not include an example where an external scalar field (such as the constrained scalar) is perturbed, but only had metric (scalar) perturbations in the exercise problems. Potentially, therefore, one might suspect that by adding an example where this is covered, one would find all of the problems solved. 

On average, the number of turns is 47, while the average number of restarts is 2, with the maximum of 4 in the sRFs. Notably, in contrast to previous cases, R2Fs, which is well-known in the literature, is now solved purely in the Jordan frame, with no introduction of another scalar at the level of action, but rather by reducing the higher derivatives on the level of perturbations, as was shown in the example.

\subsection{Context instruction}
The context ``instruction'' does not include any code, but instead describes an algorithm to compute the number of degrees of freedom for the given problems. As a result, the LLM does not analyze any of the problems by fully formulating them in Maple and running there the analysis. Instead, for some of the problems, it combined the analytical reasoning ``by hand", and a bit of Maple code, up to expanding the Ricci scalar or Ricci tensor in the action, to infer the final form of the perturbations. 

On average, the number of restarts per program is 9, with a maximum of 17 in sRFt, and a minimum of 4 in R2Fs and sRi2Ms. The number of turns is significantly higher than the rest of the cases, averaging on 67. 

Overall, its performance is significantly lower than the other cases, with only sRFt, sRMt and sRi2Ms getting a passing grade. What these problems have in common is the actions are particularly simple, as the program evaluated the tensor modes and scalars on flat background, the latter taking with vanishing background value of the scalar field. However, it should be pointed out that these problems were not solved perfectly. The program never computed the full set of the background equations of motion. In addition, in sRFt, the action given in the summary of the computation differed by a (correct) factor of $1/2$ in contrast to the action that was previously derived. In the cases of flat background, it understood this as a cosmological background, as was the case for contexts 10ex and 3broad. 
 
The largest issue among the failed problems is the decision-making. For example, in sRFs, it could never decide on the approach, and never performs any computations in Maple. In fact, the trend of doing minimal Maple computations extends to many other problems, resulting in a fail. Among them, specially, sRi2Ft reaches the 100 turns which is why its evaluation stops, while in the sRi2Fs it simply gives up after 80 turns. 

\subsection{Thinking mode: ON}\label{sec:thinking}
The previous main results are obtained with thinking turned off. In the following, we present the results when 1024 thinking tokens are allowed.
Overall, we find very little improvement from allowing thinking tokens.

\textbf{10ex | sRi2Ft}
Originally, without thinking, this problem presents an incorrect result, with the mistake taking place almost at the very end when Claude tries to reduce the order of higher-derivatives for the tensor modes. With the thinking mode ON, it makes the same mistake, again leading to the wrong result. 

\textbf{3broad | sRMt}
Claude again does the same mistake even in this problem -- flat background is interpreted as flat FLRW background instead of the Minkowski space-time, a notion which is well-known in the literature. 

\textbf{3tailored | sRFs}
In contrast to the previous run, the thinking mode manages to finish the running of the program in 58 steps, with only initial restart. However, it does exactly the same mistake as in the initial run, forgetting to perturb the constrained scalar in addition to the metric perturbations. Therefore, the result still remains a fail. 

\textbf{instruction | sRi2Fs}
Finally, with context containing the algorithm description, Claude does more elaborate computations with the thinking mode. However, it still does not manage to get to the end result. Therefore, this case also does not present a change.

\section{Conclusions}\label{sec:concl}

In their daily work, theoretical physicists encounter many problems of algorithmic character.
We explore how well a frontier LLM equipped with a CAS session can handle such problems.
Specifically, we focus on finding the degrees of freedom in modified theories of gravity and utilize completely novel problems for evaluation.

Equipped with qualitative and quantitative evaluations described before, we can now answer the questions posed in the introduction as follows:
\begin{enumerate}
  \item \textit{What are current capabilities of frontier LLMs equipped with a CAS runtime?}
    When supplied with worked examples, the LLM is able to competently use the CAS REPL and in a majority of cases solves the test problems correctly.
    We evaluate the LLM's ability on par with a first-year graduate student.
  \item \textit{How should the LLM's context be used?}
    We find that worked examples are essential for the LLM to succeed whereas an abstract description of the algorithm is not sufficient.
    Pracitioners will typically have worked examples available from their prior work; we recommend to use such examples for in-context learning.
    We find that a smaller set of example problems, chosen for representativeness, is more effective than a large set (even though in our quantitative evaluation the smaller representative set 3broad introduced a very specific failure mode of misunderstanding the problem).
    A small, targeted set can mitigate observed failure modes but may introduce new issues.
  \item \textit{What are typical failure modes and how can they be mitigated?}
    One common failure was misinterpretation of the problem (typically, because the solution of the actual problem was deemed ``trivial'' -- not unlike students in exams).
    A human in the loop would be able to correct this type of problem.
    Other mistakes such as failure to correctly apply the background equations are more worrisome but would be caught by an experienced supervisor.
    We find that tailoring the context with hand-crafted examples that cover observed failure modes can be effective in reducing mistakes.
\end{enumerate}

While the recommendations to choose the context deliberately and utilize human-AI collaboration are expected, we also make some unexpected findings.
The test problems were tailored to contain subtleties such as branching solutions.
Our expectation that their existence would be overlooked was wrong: the LLM identified most of the special cases correctly and used some of them to simplify the analysis. 
We had also expected that enabling thinking would be more useful, especially for the cases with an abstract description of the algorithm in the context.
However, the model makes little use of the thinking budget and does not improve performance.

Two aspects of the LLM's behavior are noteworthy.
First, it shows strong perseverance and goal-orientedness.
In several examples we observe its pursuing a wrong path, noticing that it is getting nowhere, and then \texttt{restart}ing the Maple session for a new attempt.
Second, the LLM shows a good ability to consider the problems abstractly.
Sometimes a problem prevents it from correctly solving the background equations of motion, at which point it proceeds with whatever it can accomplish in the rest of the computation while keeping the background solution as an abstract object.
Later it tries to go back and obtain the concrete solution.

We have only explored one class of algorithmic theoretical physics problems, with a single frontier LLM, and with a single CAS, and a fixed context.
In future work we will expand on these and especially evaluate a RAG-style setup in which the model can pull specific parts of the example calculations into context.
As we have observed strong performance on the considered test problems, we believe that CAS-equipped LLMs with in-context learning will be useful for other algorithmic computations across string theory, quantum field theory, gravity, and cosmology.

 \section*{Acknowledgements}

AH would like to thank Hugo Allaire, Elisa Ferreira, Misao Sasaki, Tadayuki Takahashi, and Ying-Li Zhang for useful discussions.
LT would like to thank Fatih Din\c{c} for advice.
AH is supported by JSPS KAKENHI Grant No. JP26K17133, and by the CD3 Google Seed grant. 
 LT is supported by JSPS under KAKENHI 24K22878 and 26K17136 and by the Royal Society under ICA\textbackslash R2\textbackslash 252140.


\bibliography{bibliography}
\bibliographystyle{icml2026}

\newpage
\appendix
\onecolumn

\section{Presentation of the problems for the context 10ex with one solved example}\label{app:context}
The full list of the problems for which we have presented detailed code with explanations in Maple is given by the following: 

\textbf{Example 1:} {Consider the w=0 Brans-Dicke theory with a linear potential on a cosmological background. Find the quadratic action for the scalar perturbation around this background, and express it in terms of the propagating modes.}

\textbf{Example 2:} { Consider a canonical scalar field theory with an arbitrary potential on a cosmological background with Einstein gravity. Find the quadratic action for the scalar perturbation around this background, and express it in terms of the propagating modes.}

\textbf{Example 3:} Consider a canonical scalar field theory with an arbitrary potential on a cosmological background with Einstein gravity. Find the quadratic action for the vector perturbation around this background, and express it in terms of the propagating modes.

\textbf{Example 4:} Consider the action consisting of Einstein gravity with cosmological constant, and a massive vector field that is non-minimally coupled to gravity through an Einstein term. Find the tensor modes, in case when the vector field has a vanishing background value, but the space-time corresponds to the de Sitter Universe.

\textbf{Example 5:} Consider the Einstein gravity in flat space-time. Find the quadratic action for the scalar perturbation around this background, and express it in terms of the propagating modes.

\textbf{Example 6:} Consider the Einstein gravity with cosmological constant in flat space-time. Find the quadratic action for the scalar perturbation around this background, and express it in terms of the propagating modes.

\textbf{Example 7:}  Consider a non-minimally coupled scalar field to the Ricci scalar, with a sin-potential. Express the action in terms of the propagating scalar modes.

\textbf{Example 8:} Consider Einstein gravity with a cosmological constant, and a scalar field with a quartic potential. Find the quadratic action for the tensor perturbations around this background, and express it in terms of the propagating modes.

\textbf{Example 9:} Consider a k-essence field theory with an arbitrary potential on a cosmological background with Einstein gravity. Find the quadratic action for the vector perturbation around this background, and express it in terms of the propagating modes.

\textbf{Example 10:} Consider a k-essence field theory with an arbitrary potential on a cosmological background with Einstein gravity. Find the quadratic action for the scalar perturbation around this background, and express it in terms of the propagating modes.

One should note that while the above formulations are essentially loose, they are supplemented with a solved step-by-step code, and thus each of them automatically comes with a defined action. As an illustration, we present the Example 2: 
\small
\begin{verbatim}
> #Example 2: Consider a canonical scalar field theory with an arbitrary
> potential on a cosmological background with Einstein gravity. Find the
> quadratic action for the scalar perturbation around this background, and
> express it in terms of the propagating modes.
> 
> #Step 1: Setup
> 
> restart
> interface(prettyprint = 0):
> #Define the background metric with scalar perturbations in the longitudinal
> gauge. Epsilon is the perturbative parameter.
> g_metric := Matrix(4, 4, [[-N(t)^2*(1 + 2*epsilon*phi(t, x)), 0, 0, 0], [0,
> a(t)^2*(1 + 2*epsilon*psi(t, x)), 0, 0], [0, 0, a(t)^2*(1 + 2*epsilon*psi(t,
> x)), 0], [0, 0, 0, a(t)^2*(1 + 2*epsilon*psi(t, x))]]):
> with(Physics):
> ds2 := ((LinearAlgebra[Transpose](<dt, dx, dy, dz>)) . g_metric) . <dt, dx,
> dy, dz>
Typesetting:-mprintslash([(ds2 := -dt^2*N(t)^2*(1+2*epsilon*phi(t,x))+dx^2*a(t)
^2*(1+2*epsilon*psi(t,x))+dy^2*a(t)^2*(1+2*epsilon*psi(t,x))+dz^2*a(t)^2*(1+2*
epsilon*psi(t,x)))],[-dt^2*N(t)^2*(1+2*epsilon*phi(t,x))+dx^2*a(t)^2*(1+2*
epsilon*psi(t,x))+dy^2*a(t)^2*(1+2*epsilon*psi(t,x))+dz^2*a(t)^2*(1+2*epsilon*
psi(t,x))])
> Setup(mathematicalnotation = true, dimension = 4, signature = `-+++`)
[dimension = 4, mathematicalnotation = true, signature = `- + + +`]
> Coordinates(Z = [t, x, y, z])
`Systems of spacetime coordinates are: {Z = (t, x, y, z)} ` {Z}
> Setup(metric = ds2)
_______________________________________________________ 
Coordinates: [t, x, y,z]. Signature: (- + + +) 
_______________________________________________________
Typesetting:-mprintslash([g_[mu,nu] = Matrix(4,4,{(1, 1) = N(t)^2*(-1-2*epsilon
*phi(t,x)), (2, 2) = a(t)^2*(1+2*epsilon*psi(t,x)), (3, 3) = a(t)^2*(1+2*
epsilon*psi(t,x)), (4, 4) = a(t)^2*(1+2*epsilon*psi(t,x))},datatype = anything,
storage = triangular[upper],order = Fortran_order,shape = [symmetric])],[g_[mu,
nu] = Matrix(%id = 36893488152155112260)])
_______________________________________________________ 
[metric = {(1, 1) =
-N(t)^2*(1+2*epsilon*phi(t,x)), (2, 2) = a(t)^2*(1+2* epsilon*psi(t,x)), (3, 3)
= a(t)^2*(1+2*epsilon*psi(t,x)), (4, 4) = a(t)^2*(1+2 *epsilon*psi(t,x))},
spaceindices = lowercaselatin_is]
> 
> #Define the scalar field:
> Sigma:=sigma(t)+epsilon*dSg(t,x)
Typesetting:-mprintslash([(Sigma := sigma(t)+epsilon*dSg(t,x))],[sigma(t)+
epsilon*dSg(t,x)])
> #Define the kinetic term for the scalar field:
> Define(redo, sigma_d[mu] = D_[mu](Sigma, [Z]), minimizetensorcomponents):
Defined objects with tensor properties
> dSig_sq := add(rhs(sigma_d[~mu, Array])[i1]*rhs(sigma_d[mu, Array])[i1], i1 =
> 1 .. 4)
Typesetting:-mprintslash([(dSig_sq := -(diff(sigma(t),t)+epsilon*diff(dSg(t,x),
t))^2/N(t)^2/(1+2*epsilon*phi(t,x))+epsilon^2/a(t)^2/(1+2*epsilon*psi(t,x))*
diff(dSg(t,x),x)^2)],[-(diff(sigma(t),t)+epsilon*diff(dSg(t,x),t))^2/N(t)^2/(1+
2*epsilon*phi(t,x))+epsilon^2/a(t)^2/(1+2*epsilon*psi(t,x))*diff(dSg(t,x),x)^2]
)
> 
> 
> #Define the action:
> det_g := LinearAlgebra[Determinant](rhs(g_[mu, nu, Array]))
Typesetting:-mprintslash([(det_g := N(t)^2*(-1-2*epsilon*phi(t,x))*a(t)^6*(1+2*
epsilon*psi(t,x))^3)],[N(t)^2*(-1-2*epsilon*phi(t,x))*a(t)^6*(1+2*epsilon*psi(t
,x))^3])
> action:=sqrt(-det_g)*(Mp^2/2*Ricci[scalar]-1/2*dSig_sq-V(Sigma)):
> action_s := simplify(action):
> Assume(0 < N(t), 0 < a(t))
{t::real, N(t)::RealRange(Open(0),infinity), a(t)::RealRange(Open(0),infinity)}
> 
> #Step 2: Background equations of motion
> 
> #To find the background equations of motion, we vary the action with respect
> to the lapse N, which we then set to 1, and the other variables from the
> ansatz that are not perturbations (multiplied with epsilon). For this we need
> the action that is at zeroth order in perturbations.
> 
> 
> background_action := simplify(series(action_s, epsilon, 1)):
> background_action1 := convert(convert(background_action, polynom), D):
> lapse_eq := expand(simplify(subs(t1 = t,
> convert(Fundiff(Intc(background_action1, t), N(t1)), D)))):
> a_eq := expand(simplify(subs(t1 = t, convert(Fundiff(Intc(background_action1,
> t), a(t1)), D)))):
> scalar_eq:= expand(simplify(subs(t1 = t,
> convert(Fundiff(Intc(background_action1, t), sigma(t1)), D)))):
> 
> 
> N := t -> 1
Typesetting:-mprintslash([(N := t -> 1)],[t -> 1])
> #Solve the background equations in terms of the variables. The best way is to
> usually solve them in terms of the highest-order derivatives.
> background_eom := simplify(solve({scalar_eq, lapse_eq, a_eq}, [D[1, 1](a)(t),
> (D@@2)(sigma)(t), V(sigma(t))])[1])
Warning, solve may be ignoring assumptions on the input variables.
Typesetting:-mprintslash([(background_eom := [(D@@2)(a)(t) = 1/2*(2*Mp^2*D(a)(t
)^2-a(t)^2*D(sigma)(t)^2)/Mp^2/a(t), (D@@2)(sigma)(t) = (-D(V)(sigma(t))*a(t)-3
*D(a)(t)*D(sigma)(t))/a(t), V(sigma(t)) = 1/2/a(t)^2*(6*Mp^2*D(a)(t)^2-a(t)^2*D
(sigma)(t)^2)])],[[(D@@2)(a)(t) = 1/2*(2*Mp^2*D(a)(t)^2-a(t)^2*D(sigma)(t)^2)/
Mp^2/a(t), (D@@2)(sigma)(t) = (-D(V)(sigma(t))*a(t)-3*D(a)(t)*D(sigma)(t))/a(t)
, V(sigma(t)) = 1/2/a(t)^2*(6*Mp^2*D(a)(t)^2-a(t)^2*D(sigma)(t)^2)]])
> #Step 3: First-order action
> 
> #Expand the action to first order in perturbations, and confirm that it
> vanishes after substituting the background equations of motion
> action_1st_order := series(action_s, epsilon, 2):
> action_1st_order1 := simplify(convert(convert(action_1st_order, polynom), D)):
> simplify(subs(background_eom, subs(tt1 = t, xx1 = t,
> convert(Fundiff(Intc(action_1st_order1, t, x), phi(tt1, xx1)), D))))
0
> simplify(subs(background_eom, subs(tt1 = t, xx1 = t,
> convert(Fundiff(Intc(action_1st_order1, t, x), psi(tt1, xx1)), D))))
0
> simplify(subs(background_eom, subs(tt1 = t, xx1 = t,
> convert(Fundiff(Intc(action_1st_order1, t, x), dSg(tt1, xx1)), D))))
0
> #Since everything is zero, we can proceed to the next step.
> 
> #Step 4: Second-order action
> 
> #Expand the action to second order in perturbations.
> 
> action_2nd_order := series(action_s, epsilon, 3):
> action_2nd_order1 := simplify(convert(coeff(convert(series(action_2nd_order,
> epsilon, 3), polynom), epsilon^2), D)):
> 
> #Simplify its form due to spherical symmetry.
> phi := unapply(ph(t)*cos(k*x)/sqrt(Pi), t, x)
Typesetting:-mprintslash([(phi := (t, x) -> 1/Pi^(1/2)*ph(t)*cos(k*x))],[(t, x)
-> 1/Pi^(1/2)*ph(t)*cos(k*x)])
> psi := unapply(ps(t)*cos(k*x)/sqrt(Pi), t, x)
Typesetting:-mprintslash([(psi := (t, x) -> 1/Pi^(1/2)*ps(t)*cos(k*x))],[(t, x)
-> 1/Pi^(1/2)*ps(t)*cos(k*x)])
> dSg := unapply(si(t)*cos(k*x)/sqrt(Pi), t, x)
Typesetting:-mprintslash([(dSg := (t, x) -> 1/Pi^(1/2)*si(t)*cos(k*x))],[(t, x)
-> 1/Pi^(1/2)*si(t)*cos(k*x)])
> action_2nd_order2 := (simplify(int(action_2nd_order1, x = -Pi .. Pi)) assuming
> k::posint):
> action_ready := (simplify(subs(background_eom, action_2nd_order2)) assuming
> k::posint):
> action_ready:
> 
> 
> #Step 5: Analyze the perturbations
> 
> #In this step, we will find the kinetic matrix of the action
> 
> fld:=[ph,ps,si]
Typesetting:-mprintslash([(fld := [ph, ps, si])],[[ph, ps, si]])
> action_scalars:=action_ready:
> #Check if there are second-order derivatives. If yes, partially integrate to
> bring it to the form: K_ij*D(field_i)(t)*D(field_j)(t)
> 
> for i from 1 to 3 do print(simplify(coeff(action_scalars, D[1,
> 1](fld[i])(t)))): end do: unassign('i'):
0 -3*Mp^2*(ph(t)-ps(t))*a(t)^3 0
> integrate_by_parts_tt:=proc(act::algebraic,var::algebraic,coo::name) local f,
> t, loc; t := coo; loc := coeff(act,D[1,1](var)(t)); f := act -
> loc*D[1,1](var)(t) - D(var)(t)*convert(diff(loc,t),D); f end proc:
> 
> action_scalars2:=simplify(subs(background_eom,
> integrate_by_parts_tt(action_scalars, ps, t))):
> for i from 1 to 3 do print(simplify(coeff(action_scalars2, D[1,
> 1](fld[i])(t)))): end do: unassign('i'):
0 0 0
> 
> #Define the kinetic matrix and check its determinant.
> KM:=Matrix(3): # Kinetic matrix
> 
> for i from 1 to 3 do for j from 1 to 3 do if i = j then KM[i, i] :=
> coeff(action_scalars2, D[1](fld)[i](t)^2); else KM[i, j] :=
> coeff(coeff(action_scalars2, D[1](fld[i])(t)), D[1](fld[j])(t)) / 2; end if;
> end do; end do;
> KM
Typesetting:-mprintslash([Matrix(3,3,{(2, 2) = -3*a(t)^3*Mp^2, (3, 3) = 1/2*a(t
)^3},datatype = anything,storage = rectangular,order = Fortran_order,shape = []
)],[Matrix(%id = 36893488152158363268)])
> LinearAlgebra[Determinant](KM)
0
> vecKM:=LinearAlgebra[LinearSolve](KM, <0, 0, 0>)
Typesetting:-mprintslash([(vecKM := Vector[column](3,{1 = _t[1]},datatype =
anything,storage = rectangular,order = Fortran_order,shape = []))],[Vector[
column](%id = 36893488152158365444)])
> #If the determinant of the kinetic matrix is non-zero, stop. You have found
> the final expression. Otherwise, check if the components of vecKM have some
> dependence on the background variables. If yes, make a substitution such that
> this is removed. If no, check the non-zero variable, and integrate out this
> field in the next step.
> 
> #Step 6: Integrate out ph
> 
> #Check if the action has terms linear in derivatives of ph. If yes, then
> partially integrate.
> simplify(coeff(action_scalars2, D(ph)(t)))
9*Mp^2*(ph(t)-ps(t))*a(t)^2*D(a)(t)
> 
> integrate_by_parts_Dt_t:=proc(act::algebraic,var::algebraic,coo::name) local
> f, t, loc; t := coo; loc := coeff(coeff(act,D(var)(t)),var(t));
> f:=act-loc*D(var)(t)*var(t)-1/2*convert(diff(loc,t),D)*var(t)^2; f end proc:
> 
> integrate_by_parts_t:=proc(act::algebraic,var::algebraic,coo::name) local f,
> t, loc; t := coo; loc := coeff(act,D(var)(t)); f := act - loc*D(var)(t) -
> var(t)*convert(diff(loc,t),D); f end proc:
> 
> action_scalars3:=simplify(subs(background_eom,
> integrate_by_parts_Dt_t(action_scalars2, ph, t))):
> action_scalars4:=simplify(subs(background_eom,
> integrate_by_parts_t(action_scalars3, ph, t))):
> 
> #Find the constraint for ph
> 
> 
> find_constraint:=proc(act::algebraic,var::algebraic,coo::name) local f, t, t1;
> t := coo;
> f:=subs(t1=t,convert(Physics[Fundiff](Physics[Intc](act,t),var(t1)),D)): f end
> proc:
> 
> ph_constraint := simplify(find_constraint(action_scalars4, ph, t)):
> 
> #If the constraint can be solved for ph, solve it, and replace the value of ph
> with this solution. Otherwise, solve it for the field for which you can solve
> it, and replace that field.
> 
> ph_solution := simplify(solve(ph_constraint, ph(t))):
Warning, solve may be ignoring assumptions on the input variables.
> ph := unapply(ph_solution, t)
Typesetting:-mprintslash([(ph := t -> (-6*Mp^2*a(t)*D(a)(t)*D(ps)(t)-2*ps(t)*Mp
^2*k^2+a(t)^2*si(t)*D(V)(sigma(t))+a(t)^2*D(si)(t)*D(sigma)(t))/(-6*Mp^2*D(a)(t
)^2+a(t)^2*D(sigma)(t)^2))],[t -> (-6*Mp^2*a(t)*D(a)(t)*D(ps)(t)-2*ps(t)*Mp^2*k
^2+a(t)^2*si(t)*D(V)(sigma(t))+a(t)^2*D(si)(t)*D(sigma)(t))/(-6*Mp^2*D(a)(t)^2+
a(t)^2*D(sigma)(t)^2)])
> 
> #Step 7: Check the resulting action and repeat the procedeure of step 5
> 
> action_scalars5:=simplify(subs(background_eom, action_scalars4)):
> fld:=[ps,si]
Typesetting:-mprintslash([(fld := [ps, si])],[[ps, si]])
> 
> for i from 1 to 2 do print(simplify(coeff(action_scalars5, D[1,
> 1](fld[i])(t)^2))): end do: unassign('i'):
0 0
> 
> #Kinetic matrix
> KM:=Matrix(2): # Kinetic matrix
> 
> for i from 1 to 2 do for j from 1 to 2 do if i = j then KM[i, i] :=
> coeff(action_scalars5, D[1](fld)[i](t)^2); else KM[i, j] :=
> coeff(coeff(action_scalars5, D[1](fld[i])(t)), D[1](fld[j])(t)) / 2; end if;
> end do; end do;
> KM
Typesetting:-mprintslash([Matrix(2,2,{(1, 1) = -3*D(sigma)(t)^2*Mp^2*a(t)^5/(-6
*Mp^2*D(a)(t)^2+a(t)^2*D(sigma)(t)^2), (1, 2) = 3*Mp^2*D(sigma)(t)*D(a)(t)*a(t)
^4/(-6*Mp^2*D(a)(t)^2+a(t)^2*D(sigma)(t)^2), (2, 1) = 3*Mp^2*D(sigma)(t)*D(a)(t
)*a(t)^4/(-6*Mp^2*D(a)(t)^2+a(t)^2*D(sigma)(t)^2), (2, 2) = -3*Mp^2*D(a)(t)^2*a
(t)^3/(-6*Mp^2*D(a)(t)^2+a(t)^2*D(sigma)(t)^2)},datatype = anything,storage =
rectangular,order = Fortran_order,shape = [])],[Matrix(%id =
36893488152227830828)])
> LinearAlgebra[Determinant](KM)
0
> vecKM:=LinearAlgebra[LinearSolve](KM, <0, 0>)
Typesetting:-mprintslash([(vecKM := Vector[column](2,{1 = _t0[1], 2 = 1/D(a)(t)
*a(t)*D(sigma)(t)*_t0[1]},datatype = anything,storage = rectangular,order =
Fortran_order,shape = []))],[Vector[column](%id = 36893488152254155700)])
> #Make substitution so that vecKM components do not depend on the background
> 
> si:=unapply(si2(t)+a(t)*D(sigma)(t)*1/D(a)(t)*ps(t),t)
Typesetting:-mprintslash([(si := t -> si2(t)+a(t)*D(sigma)(t)/D(a)(t)*ps(t))],[
t -> si2(t)+a(t)*D(sigma)(t)/D(a)(t)*ps(t)])
> action_scalars6:=simplify(subs(background_eom, action_scalars5)):
> 
> fld:=[ps,si2]
Typesetting:-mprintslash([(fld := [ps, si2])],[[ps, si2]])
> for i from 1 to 2 do print(simplify(coeff(action_scalars6, D[1,
> 1](fld[i])(t)^2))): end do: unassign('i'):
0 0
> 
> #Kinetic matrix
> KM:=Matrix(2): # Kinetic matrix
> 
> for i from 1 to 2 do for j from 1 to 2 do if i = j then KM[i, i] :=
> coeff(action_scalars6, D[1](fld)[i](t)^2); else KM[i, j] :=
> coeff(coeff(action_scalars6, D[1](fld[i])(t)), D[1](fld[j])(t)) / 2; end if;
> end do; end do;
> KM
Typesetting:-mprintslash([Matrix(2,2,{(2, 2) = -3*Mp^2*D(a)(t)^2*a(t)^3/(-6*Mp^
2*D(a)(t)^2+a(t)^2*D(sigma)(t)^2)},datatype = anything,storage = rectangular,
order = Fortran_order,shape = [])],[Matrix(%id = 36893488152231181836)])
> LinearAlgebra[Determinant](KM)
0
> vecKM:=LinearAlgebra[LinearSolve](KM, <0, 0>)
Typesetting:-mprintslash([(vecKM := Vector[column](2,{1 = _t1[1]},datatype =
anything,storage = rectangular,order = Fortran_order,shape = []))],[Vector[
column](%id = 36893488152231174980)])
> 
> #Step 8: Integrate out ps
> 
> #Check if the action has terms linear in derivatives of ps. If yes, then
> partially integrate.
> simplify(coeff(action_scalars6, D(ps)(t)))
(3*Mp^2*ps(t)*D(a)(t)^3+2*Mp^2*ps(t)*D(a)(t)*k^2-a(t)^3*ps(t)*D(sigma)(t)*D(V)(
sigma(t))+3*a(t)^2*ps(t)*D(a)(t)*D(sigma)(t)^2-si2(t)*a(t)^2*D(a)(t)*D(V)(sigma
(t)))*a(t)^2/D(a)(t)^2
> 
> action_scalars7:=simplify(subs(background_eom,
> integrate_by_parts_Dt_t(action_scalars6, ps, t))):
> action_scalars8:=simplify(subs(background_eom,
> integrate_by_parts_t(action_scalars7, ps, t))):
> 
> 
> ps_constraint := simplify(find_constraint(action_scalars8, ps, t)):
> ps_solution := simplify(solve(ps_constraint, ps(t))):
Warning, solve may be ignoring assumptions on the input variables.
> ps := unapply(ps_solution, t)
Typesetting:-mprintslash([(ps := t -> 1/4*a(t)*(2*Mp^2*si2(t)*a(t)*D(a)(t)*D(V)
(sigma(t))+6*Mp^2*si2(t)*D(a)(t)^2*D(sigma)(t)+2*Mp^2*D(si2)(t)*a(t)*D(a)(t)*D(
sigma)(t)-si2(t)*a(t)^2*D(sigma)(t)^3)/Mp^4/D(a)(t)/k^2)],[t -> 1/4*a(t)*(2*Mp^
2*si2(t)*a(t)*D(a)(t)*D(V)(sigma(t))+6*Mp^2*si2(t)*D(a)(t)^2*D(sigma)(t)+2*Mp^2
*D(si2)(t)*a(t)*D(a)(t)*D(sigma)(t)-si2(t)*a(t)^2*D(sigma)(t)^3)/Mp^4/D(a)(t)/k
^2])
> 
> 
> #Step 9: Check the resulting action and repeat the procedeure of step 5
> 
> action_scalars9:=simplify(subs(background_eom, action_scalars8)):
> fld:=[si2]
Typesetting:-mprintslash([(fld := [si2])],[[si2]])
> 
> for i from 1 to 1 do print(simplify(coeff(action_scalars9, D[1,
> 1](fld[i])(t)^2))): end do: unassign('i'):
0
> 
> #Kinetic matrix
> KM:=Matrix(1): # Kinetic matrix
> 
> for i from 1 to 1 do for j from 1 to 1 do if i = j then KM[i, i] :=
> coeff(action_scalars9, D[1](fld)[i](t)^2); else KM[i, j] :=
> coeff(coeff(action_scalars9, D[1](fld[i])(t)), D[1](fld[j])(t)) / 2; end if;
> end do; end do;
> KM
Typesetting:-mprintslash([Matrix(1,1,{(1, 1) = 1/2*a(t)^3},datatype = anything,
storage = rectangular,order = Fortran_order,shape = [])],[Matrix(%id =
36893488152238293108)])
> LinearAlgebra[Determinant](KM)
1/2*a(t)^3
> #Determinant is not vanishing, and therefore you are done. (Assuming that
> there are no higher-order derivatives overall)
> 
> 
> collect(action_scalars9, [D(si2)(t)^2,D(si2)(t),si2(t)],simplify)
> 
1/2*a(t)^3*D(si2)(t)^2-1/2*a(t)^4/Mp^2*si2(t)/D(a)(t)*D(sigma)(t)^2*D(si2)(t)-1
/8*a(t)*(4*Mp^4*a(t)^2*(D@@2)(V)(sigma(t))*D(a)(t)^2+4*Mp^4*D(a)(t)^2*k^2+4*Mp^
2*a(t)^3*D(a)(t)*D(sigma)(t)*D(V)(sigma(t))+6*Mp^2*a(t)^2*D(a)(t)^2*D(sigma)(t)
^2-a(t)^4*D(sigma)(t)^4)/D(a)(t)^2/Mp^4*si2(t)^2

\end{verbatim}

\section{The recipe for the context instructions}\label{app:instruction}

Work-out the perturbations based on the following the general instructions: 

The following is VERY IMPORTANT: SET interface(prettyprint = 0):

(A) Specify the background
One of the main reasons to study any theory is to describe physical phenomena, such as, for example, the acceleration of the Universe, black-holes, or cosmological defects. These are the solutions to the background equations of motion, found by varying the action with respect to fields. Notably, depending on the background, the number of degrees of freedom (dof) can change. Therefore, when performing the analysis to calculate the dof, it is best to assume that the background is always satisfied. One can ensure this by simply assuming that the background equations of motion always hold at any step in the computation.
(B) Prepare the perturbations
Once you specify the background (which can also be trivial), it’s time to perturb around it. These perturbations will provide the notion of the degrees of freedom. To find them, there are several preparatory things to consider before studying their corresponding action. In particular, it is useful to answer the following questions to decide on the best way forward:
(i) Does your system have any symmetries?
Here, by symmetries, we mean what is left after one separates the time and space components. If
then the answer is yes, this might suggest the form of your perturbations, from a general study of components of your field to a much simpler form. For example, in space-times that are homogeneous and isotropic, it is possible to do a decomposition of the modes according to the group of spatial rotations, and decompose the fields into scalars, vectors, and tensors. If the space-time has broken anisotropies in one direction and three spatial dimensions, one can instead decompose the components of the fields into even and odd modes.

(B) Does your system have any gauge redundancy?
If yes, analyzing first how the perturbations are transforming under gauge redundancy could be very beneficial: if one fixes the gauge such that it matches the gauge invariant variables, it can decrease the length of the computation. Other gauges are possible too, as long as one makes sure that the gauge is fully fixed.
It should be stressed that the above two questions can be avoided – it is perfectly fine if one considers all components of the fields, and does not even fix the gauge. These two decisions would lead to equivalent results, but, also can make the computation significantly longer.

(C) Expand the action
Once one decides on the background and the type of perturbations around it, it is time to study the perturbations themselves. This can be analyzed either on the level of action or directly on the level of equations of motion. However, studying them directly from the action is more beneficial, since one can then have a clear view into the conditions under which the degrees of freedom are well behaved, such as
the positivity of the factors multiplying their kinetic terms, necessary to avoid the ghost modes.

In general, one is supposed to expand the action to the required order in perturbations. If there is a non-vanishing background for the fields, the leading order corresponds to the equations of motion, which are subsequently satisfied. Thus, at first order, the corrections vanish. The kinetic terms for the perturbations are then at second order in the action, and one can also study higher orders to analyze the non-linear corrections. Note that this also holds if the background is trivial, eg, including vanishing background values of the fields, together with the Minkowski space-time.

(D) Analyze the theory
Once the action is ready, it is time to analyze the modes. Let us first assume that there are at most two time derivatives acting on the fields, and that all constraints are holonomic. Then, essentially, the procedure to find the propagating modes is given by the following:
(i) Locate the fields that are not propagating – the constrained modes.
(ii) Find their corresponding constraint by varying the action with respect to them.
(iii) Solve it, and substitute it back into the action. Note that in the case of non-linearities, this is usually
done perturbatively.
(iv) Check if the determinant of the kinetic matrix for the Lagrangian density of fields is vanishing. If yes, this means that there are more non-propagating fields. Locate them in the
Lagrangian density. If all of them appear with the form where they carry terms quadratic in time-derivatives, this means that one should perform
a substitution, which will render one of them to be non-propagating, and then one can repeat the
procedure (i) – (iv).
(v) If, after repeating the procedure, the determinant of the kinetic matrix is not vanishing, and your system contains no higher derivatives, then you have found the Lagrangian density in terms of the propagating modes. 

(E) Higher-order derivatives
The above procedure works for a wide range of theories. However, one can also easily encounter theories that contain higher-order derivatives. Luckily, this does not drastically complicate the analysis. The key to analyzing it is to reduce the order of derivatives until one reaches maximally two time derivatives per term in the Lagrangian. 

\section{Claude output example}\label{app:example}
In the following, we present as an example, the Claude output for the sRMt, with the context 10ex. 

\small
\begin{verbatim}

# +++++++++++++++++ META TRY 1 ++++++++++++++ 

# ----------------- TURN 1 -----------------


# I'll solve this step-by-step, following the pattern from the example
# problems.

# ## Step 1: Setup
restart;
interface(prettyprint = 0):

# Define the background metric with tensor perturbations. Epsilon is the perturbative parameter.
g_metric := Matrix(4, 4, [[-N(t)^2, 0,0, 0], [0, a(t)^2, 0, 0], [0, 0,
1/2*a(t)^2*(epsilon*sqrt(2)*tenp(t, x) + 2),
1/2*epsilon*a(t)^2*sqrt(2)*tenc(t, x)], [0, 0,
1/2*epsilon*a(t)^2*sqrt(2)*tenc(t, x), -1/2*a(t)^2*(epsilon*sqrt(2)*tenp(t, x)
- 2)]]):

with(Physics):
ds2 := ((LinearAlgebra[Transpose](<dt, dx, dy, dz>)) . g_metric) . <dt, dx, dy, dz>;
~ ds2 := -dt^2*N(t)^2+dx^2*a(t)^2+dy*(1/2*a(t)^2*(epsilon*2^(1/2)*tenp(t,x)+2)*dy
~ +1/2*epsilon*a(t)^2*2^(1/2)*tenc(t,x)*dz)+dz*(1/2*epsilon*a(t)^2*2^(1/2)*tenc(t
~ ,x)*dy-1/2*a(t)^2*(epsilon*2^(1/2)*tenp(t,x)-2)*dz)
# ----------------- TURN 2 -----------------
Setup(mathematicalnotation = true, dimension = 4, signature = `-+++`);
Coordinates(Z = [t, x, y, z]);
Setup(metric = ds2);
~ [dimension = 4, mathematicalnotation = true, signature = `- + + +`]
~ `Systems of spacetime coordinates are: {Z = (t, x, y, z)} `
~ {Z}
~ "_______________________________________________________"
~ "Coordinates: [t, x, y, z]. Signature: (- + + +)"
~ "_______________________________________________________"
~ g_[mu,nu] = Matrix(4,4,{(1, 1) = -N(t)^2, (2, 2) = a(t)^2, (3, 3) = 1/2*a(t)^2*
~ (epsilon*2^(1/2)*tenp(t,x)+2), (3, 4) = 1/2*epsilon*a(t)^2*2^(1/2)*tenc(t,x), (
~ 4, 4) = -1/2*a(t)^2*(epsilon*2^(1/2)*tenp(t,x)-2)},datatype = anything,storage
~ = triangular[upper],order = Fortran_order,shape = [symmetric])
~ "_______________________________________________________"
~ [metric = {(1, 1) = -N(t)^2, (2, 2) = a(t)^2, (3, 3) = 1/2*a(t)^2*(epsilon*2^(1
~ /2)*tenp(t,x)+2), (3, 4) = 1/2*epsilon*a(t)^2*2^(1/2)*tenc(t,x), (4, 4) = -1/2*
~ a(t)^2*(epsilon*2^(1/2)*tenp(t,x)-2)}, spaceindices = lowercaselatin_is]
# ----------------- TURN 3 -----------------
# Define the scalar field (background only for tensor perturbations):
Sigma := sigma(t);

# Define the action: Mp^2/2*R + 1/2*sigma^2*(m^2 + f(R))
det_g := LinearAlgebra[Determinant](rhs(g_[mu, nu, Array]));

action := sqrt(-det_g)*(Mp^2/2*Ricci[scalar] + 1/2*Sigma^2*(m^2 + f(Ricci[scalar])));
action_s := simplify(action):
Assume(0 < N(t), 0 < a(t));
~ Sigma := sigma(t)
~ det_g := 1/2*N(t)^2*a(t)^6*(tenp(t,x)^2*epsilon^2+epsilon^2*tenc(t,x)^2-2)
~ action := 1/2*(-2*N(t)^2*a(t)^6*(tenp(t,x)^2*epsilon^2+epsilon^2*tenc(t,x)^2-2)
~ )^(1/2)*(1/4*Mp^2*(16*diff(tenp(t,x),t)*tenc(t,x)^2*tenp(t,x)*diff(a(t),t)*a(t)
~ *N(t)*epsilon^4-8*diff(tenp(t,x),t)*tenc(t,x)*diff(tenc(t,x),t)*tenp(t,x)*a(t)^
~ 2*N(t)*epsilon^4+16*tenc(t,x)*diff(tenc(t,x),t)*tenp(t,x)^2*diff(a(t),t)*a(t)*N
~ (t)*epsilon^4+48*diff(a(t),t)^2*N(t)+48*tenc(t,x)^2*diff(a(t),t)*a(t)*diff(N(t)
~ ,t)*epsilon^2+48*tenp(t,x)^2*diff(a(t),t)*a(t)*diff(N(t),t)*epsilon^2+8*diff(
~ tenp(t,x),t)*tenp(t,x)*a(t)^2*diff(N(t),t)*epsilon^2+8*tenc(t,x)*diff(tenc(t,x)
~ ,t)*a(t)^2*diff(N(t),t)*epsilon^2+3*diff(tenp(t,x),t)^2*tenc(t,x)^2*a(t)^2*N(t)
~ *epsilon^4-4*diff(tenp(t,x),t)*tenp(t,x)^3*a(t)^2*diff(N(t),t)*epsilon^4-12*
~ tenc(t,x)^4*diff(a(t),t)*a(t)*diff(N(t),t)*epsilon^4-4*tenc(t,x)^3*diff(tenc(t,
~ x),t)*a(t)^2*diff(N(t),t)*epsilon^4+24*tenc(t,x)^2*tenp(t,x)^2*diff(a(t),t)^2*N
~ (t)*epsilon^4+3*diff(tenc(t,x),t)^2*tenp(t,x)^2*a(t)^2*N(t)*epsilon^4-12*tenp(t
~ ,x)^4*diff(a(t),t)*a(t)*diff(N(t),t)*epsilon^4-32*diff(tenp(t,x),t)*tenp(t,x)*
~ diff(a(t),t)*a(t)*N(t)*epsilon^2-32*tenc(t,x)*diff(tenc(t,x),t)*diff(a(t),t)*a(
~ t)*N(t)*epsilon^2-4*diff(tenp(t,x),t)*tenc(t,x)^2*tenp(t,x)*a(t)^2*diff(N(t),t)
~ *epsilon^4+16*diff(tenp(t,x),t)*tenp(t,x)^3*diff(a(t),t)*a(t)*N(t)*epsilon^4+16
~ *tenc(t,x)^3*diff(tenc(t,x),t)*diff(a(t),t)*a(t)*N(t)*epsilon^4-24*tenc(t,x)^2*
~ tenp(t,x)^2*diff(a(t),t)*a(t)*diff(N(t),t)*epsilon^4-4*tenc(t,x)*diff(tenc(t,x)
~ ,t)*tenp(t,x)^2*a(t)^2*diff(N(t),t)*epsilon^4+8*tenc(t,x)*tenp(t,x)*diff(tenp(t
~ ,x),x)*diff(tenc(t,x),x)*N(t)^3*epsilon^4+48*a(t)*N(t)*diff(a(t),t $ 2)+tenc(t,
~ x)^2*diff(tenc(t,x),x)^2*N(t)^3*epsilon^4+tenp(t,x)^2*diff(tenp(t,x),x)^2*N(t)^
~ 3*epsilon^4+12*tenc(t,x)^4*diff(a(t),t)^2*N(t)*epsilon^4-3*tenc(t,x)^2*diff(
~ tenp(t,x),x)^2*N(t)^3*epsilon^4+12*tenp(t,x)^4*diff(a(t),t)^2*N(t)*epsilon^4-3*
~ tenp(t,x)^2*diff(tenc(t,x),x)^2*N(t)^3*epsilon^4-6*diff(tenp(t,x),t)^2*a(t)^2*N
~ (t)*epsilon^2-48*tenc(t,x)^2*diff(a(t),t)^2*N(t)*epsilon^2-6*diff(tenc(t,x),t)^
~ 2*a(t)^2*N(t)*epsilon^2-48*tenp(t,x)^2*diff(a(t),t)^2*N(t)*epsilon^2+6*diff(
~ tenp(t,x),x)^2*N(t)^3*epsilon^2+6*diff(tenc(t,x),x)^2*N(t)^3*epsilon^2-48*diff(
~ a(t),t)*a(t)*diff(N(t),t)+24*tenc(t,x)^2*tenp(t,x)^2*a(t)*N(t)*diff(a(t),t $ 2)
~ *epsilon^4+4*tenc(t,x)^2*tenp(t,x)*a(t)^2*N(t)*diff(tenp(t,x),t $ 2)*epsilon^4+
~ 4*tenc(t,x)*tenp(t,x)^2*a(t)^2*N(t)*diff(tenc(t,x),t $ 2)*epsilon^4-8*tenp(t,x)
~ *a(t)^2*N(t)*diff(tenp(t,x),t $ 2)*epsilon^2-48*tenc(t,x)^2*a(t)*N(t)*diff(a(t)
~ ,t $ 2)*epsilon^2-48*tenp(t,x)^2*a(t)*N(t)*diff(a(t),t $ 2)*epsilon^2-8*tenc(t,
~ x)*a(t)^2*N(t)*diff(tenc(t,x),t $ 2)*epsilon^2+12*tenc(t,x)^4*a(t)*N(t)*diff(a(
~ t),t $ 2)*epsilon^4+4*tenc(t,x)^3*a(t)^2*N(t)*diff(tenc(t,x),t $ 2)*epsilon^4-4
~ *tenc(t,x)^2*tenp(t,x)*N(t)^3*diff(tenp(t,x),x $ 2)*epsilon^4-4*tenc(t,x)*tenp(
~ t,x)^2*N(t)^3*diff(tenc(t,x),x $ 2)*epsilon^4+12*tenp(t,x)^4*a(t)*N(t)*diff(a(t
~ ),t $ 2)*epsilon^4+4*tenp(t,x)^3*a(t)^2*N(t)*diff(tenp(t,x),t $ 2)*epsilon^4-a(
~ t)^2*diff(tenp(t,x),t)^2*tenp(t,x)^2*N(t)*epsilon^4-a(t)^2*tenc(t,x)^2*N(t)*
~ diff(tenc(t,x),t)^2*epsilon^4-4*tenc(t,x)^3*N(t)^3*diff(tenc(t,x),x $ 2)*
~ epsilon^4-4*tenp(t,x)^3*N(t)^3*diff(tenp(t,x),x $ 2)*epsilon^4+8*tenc(t,x)*N(t)
~ ^3*diff(tenc(t,x),x $ 2)*epsilon^2+8*tenp(t,x)*N(t)^3*diff(tenp(t,x),x $ 2)*
~ epsilon^2)/a(t)^2/N(t)^3/(tenp(t,x)^2*epsilon^2+epsilon^2*tenc(t,x)^2-2)^2+1/2*
~ sigma(t)^2*(m^2+f(1/2*(16*diff(tenp(t,x),t)*tenc(t,x)^2*tenp(t,x)*diff(a(t),t)*
~ a(t)*N(t)*epsilon^4-8*diff(tenp(t,x),t)*tenc(t,x)*diff(tenc(t,x),t)*tenp(t,x)*a
~ (t)^2*N(t)*epsilon^4+16*tenc(t,x)*diff(tenc(t,x),t)*tenp(t,x)^2*diff(a(t),t)*a(
~ t)*N(t)*epsilon^4+48*diff(a(t),t)^2*N(t)+48*tenc(t,x)^2*diff(a(t),t)*a(t)*diff(
~ N(t),t)*epsilon^2+48*tenp(t,x)^2*diff(a(t),t)*a(t)*diff(N(t),t)*epsilon^2+8*
~ diff(tenp(t,x),t)*tenp(t,x)*a(t)^2*diff(N(t),t)*epsilon^2+8*tenc(t,x)*diff(tenc
~ (t,x),t)*a(t)^2*diff(N(t),t)*epsilon^2+3*diff(tenp(t,x),t)^2*tenc(t,x)^2*a(t)^2
~ *N(t)*epsilon^4-4*diff(tenp(t,x),t)*tenp(t,x)^3*a(t)^2*diff(N(t),t)*epsilon^4-\
~ 12*tenc(t,x)^4*diff(a(t),t)*a(t)*diff(N(t),t)*epsilon^4-4*tenc(t,x)^3*diff(tenc
~ (t,x),t)*a(t)^2*diff(N(t),t)*epsilon^4+24*tenc(t,x)^2*tenp(t,x)^2*diff(a(t),t)^
~ 2*N(t)*epsilon^4+3*diff(tenc(t,x),t)^2*tenp(t,x)^2*a(t)^2*N(t)*epsilon^4-12*
~ tenp(t,x)^4*diff(a(t),t)*a(t)*diff(N(t),t)*epsilon^4-32*diff(tenp(t,x),t)*tenp(
~ t,x)*diff(a(t),t)*a(t)*N(t)*epsilon^2-32*tenc(t,x)*diff(tenc(t,x),t)*diff(a(t),
~ t)*a(t)*N(t)*epsilon^2-4*diff(tenp(t,x),t)*tenc(t,x)^2*tenp(t,x)*a(t)^2*diff(N(
~ t),t)*epsilon^4+16*diff(tenp(t,x),t)*tenp(t,x)^3*diff(a(t),t)*a(t)*N(t)*epsilon
~ ^4+16*tenc(t,x)^3*diff(tenc(t,x),t)*diff(a(t),t)*a(t)*N(t)*epsilon^4-24*tenc(t,
~ x)^2*tenp(t,x)^2*diff(a(t),t)*a(t)*diff(N(t),t)*epsilon^4-4*tenc(t,x)*diff(tenc
~ (t,x),t)*tenp(t,x)^2*a(t)^2*diff(N(t),t)*epsilon^4+8*tenc(t,x)*tenp(t,x)*diff(
~ tenp(t,x),x)*diff(tenc(t,x),x)*N(t)^3*epsilon^4+48*a(t)*N(t)*diff(a(t),t $ 2)+
~ tenc(t,x)^2*diff(tenc(t,x),x)^2*N(t)^3*epsilon^4+tenp(t,x)^2*diff(tenp(t,x),x)^
~ 2*N(t)^3*epsilon^4+12*tenc(t,x)^4*diff(a(t),t)^2*N(t)*epsilon^4-3*tenc(t,x)^2*
~ diff(tenp(t,x),x)^2*N(t)^3*epsilon^4+12*tenp(t,x)^4*diff(a(t),t)^2*N(t)*epsilon
~ ^4-3*tenp(t,x)^2*diff(tenc(t,x),x)^2*N(t)^3*epsilon^4-6*diff(tenp(t,x),t)^2*a(t
~ )^2*N(t)*epsilon^2-48*tenc(t,x)^2*diff(a(t),t)^2*N(t)*epsilon^2-6*diff(tenc(t,x
~ ),t)^2*a(t)^2*N(t)*epsilon^2-48*tenp(t,x)^2*diff(a(t),t)^2*N(t)*epsilon^2+6*
~ diff(tenp(t,x),x)^2*N(t)^3*epsilon^2+6*diff(tenc(t,x),x)^2*N(t)^3*epsilon^2-48*
~ diff(a(t),t)*a(t)*diff(N(t),t)+24*tenc(t,x)^2*tenp(t,x)^2*a(t)*N(t)*diff(a(t),t
~ $ 2)*epsilon^4+4*tenc(t,x)^2*tenp(t,x)*a(t)^2*N(t)*diff(tenp(t,x),t $ 2)*
~ epsilon^4+4*tenc(t,x)*tenp(t,x)^2*a(t)^2*N(t)*diff(tenc(t,x),t $ 2)*epsilon^4-8
~ *tenp(t,x)*a(t)^2*N(t)*diff(tenp(t,x),t $ 2)*epsilon^2-48*tenc(t,x)^2*a(t)*N(t)
~ *diff(a(t),t $ 2)*epsilon^2-48*tenp(t,x)^2*a(t)*N(t)*diff(a(t),t $ 2)*epsilon^2
~ -8*tenc(t,x)*a(t)^2*N(t)*diff(tenc(t,x),t $ 2)*epsilon^2+12*tenc(t,x)^4*a(t)*N(
~ t)*diff(a(t),t $ 2)*epsilon^4+4*tenc(t,x)^3*a(t)^2*N(t)*diff(tenc(t,x),t $ 2)*
~ epsilon^4-4*tenc(t,x)^2*tenp(t,x)*N(t)^3*diff(tenp(t,x),x $ 2)*epsilon^4-4*tenc
~ (t,x)*tenp(t,x)^2*N(t)^3*diff(tenc(t,x),x $ 2)*epsilon^4+12*tenp(t,x)^4*a(t)*N(
~ t)*diff(a(t),t $ 2)*epsilon^4+4*tenp(t,x)^3*a(t)^2*N(t)*diff(tenp(t,x),t $ 2)*
~ epsilon^4-a(t)^2*diff(tenp(t,x),t)^2*tenp(t,x)^2*N(t)*epsilon^4-a(t)^2*tenc(t,x
~ )^2*N(t)*diff(tenc(t,x),t)^2*epsilon^4-4*tenc(t,x)^3*N(t)^3*diff(tenc(t,x),x $
~ 2)*epsilon^4-4*tenp(t,x)^3*N(t)^3*diff(tenp(t,x),x $ 2)*epsilon^4+8*tenc(t,x)*N
~ (t)^3*diff(tenc(t,x),x $ 2)*epsilon^2+8*tenp(t,x)*N(t)^3*diff(tenp(t,x),x $ 2)*
~ epsilon^2)/a(t)^2/N(t)^3/(tenp(t,x)^2*epsilon^2+epsilon^2*tenc(t,x)^2-2)^2)))
~ {t::real, N(t)::RealRange(Open(0),infinity), a(t)::RealRange(Open(0),infinity)}
# ----------------- TURN 4 -----------------
# ## Step 2: Background equations of motion
background_action := simplify(series(action_s, epsilon, 1)):
background_action1 := convert(convert(background_action, polynom), D):
lapse_eq := expand(simplify(subs(t1 = t,
convert(Fundiff(Intc(background_action1, t), N(t1)), D)))):
a_eq := expand(simplify(subs(t1 = t, convert(Fundiff(Intc(background_action1,
t), a(t1)), D)))):
scalar_eq := expand(simplify(subs(t1 = t, convert(Fundiff(Intc(background_action1,
t), sigma(t1)), D)))):
# ----------------- TURN 5 -----------------
N := t -> 1;
~ N := t -> 1
# ----------------- TURN 6 -----------------
# Now for this problem, the background is flat space-time (a(t)=1), so
# let me check if that's consistent:
a := t -> 1;
# Check background equations
simplify(lapse_eq);
simplify(a_eq);
simplify(scalar_eq);
~ a := t -> 1
~ 1/2*sigma(t)^2*(m^2+f(0))
~ 1/2*(3*m^2+3*f(0))*sigma(t)^2+6*(D@@2)(sigma)(t)*D(f)(0)*sigma(t)+6*D(f)(0)*D(
~ sigma)(t)^2
~ sigma(t)*(m^2+f(0))
# ----------------- TURN 7 -----------------
# For flat background (a=1, N=1), the Ricci scalar R=0.
# The lapse equation gives: 1/2*sigma(t)^2*(m^2+f(0)) = 0
# The scalar equation gives: sigma(t)*(m^2+f(0)) = 0
# These are satisfied if either sigma(t)=0 or m^2+f(0)=0.
# Let's take sigma(t) as constant sigma0, and m^2 + f(0) = 0.
# Actually, let's check if sigma=0 works for a_eq too:
simplify(subs(sigma(t)=0, D(sigma)(t)=0, (D@@2)(sigma)(t)=0, a_eq));
~ 0
# ----------------- TURN 8 -----------------
# sigma=0 satisfies all background equations on flat background. Let's use this.
# But wait - if sigma=0, then the sigma^2 term vanishes and we just have GR at background level.
# However, sigma can have a background value. Let me reconsider.
# For sigma(t) nonzero, we need m^2 + f(0) = 0, which means m^2 = -f(0).
# And the a_eq becomes: 3*sigma^2*(m^2+f(0)) + 6*sigma''*D(f)(0)*sigma + 6*D(f)(0)*sigma'^2
# With m^2+f(0)=0: 6*D(f)(0)*(sigma*sigma'' + sigma'^2) = 6*D(f)(0)*d/dt(sigma*sigma') = 0
# This is getting complex. Let's try sigma=0 as the simplest flat background.

# Actually, looking at the problem more carefully - it says "flat background"
# With sigma=0, the action at quadratic order should still be interesting 
# because the tensor modes couple to the metric perturbations.

# Let me proceed with sigma=0 background
sigma := t -> 0;
# Re-verify
print("lapse:", simplify(lapse_eq));
print("a_eq:", simplify(a_eq));
~ sigma := t -> 0
~ "lapse:", 0
~ "a_eq:", 0
# ----------------- TURN 9 -----------------
# Good - flat background with sigma=0 is a solution.
# Now proceed to first-order and second-order action.

# Step 3: First-order action
action_1st_order := series(action_s, epsilon, 2):
action_1st_order1 := simplify(convert(convert(action_1st_order, polynom), D)):
simplify(subs(tt1 = t, xx1 = t, convert(Fundiff(Intc(action_1st_order1, t, x), tenp(tt1, xx1)), D)));
simplify(subs(tt1 = t, xx1 = t, convert(Fundiff(Intc(action_1st_order1, t, x), tenc(tt1, xx1)), D)));
~ 0
~ 0
# ----------------- TURN 10 -----------------
# Good, first-order action vanishes. Now Step 4:
# Step 4: Second-order action
action_2nd_order := series(action_s, epsilon, 3):
action_2nd_order1 := simplify(convert(coeff(convert(series(action_2nd_order,
epsilon, 3), polynom), epsilon^2), D)):

# Simplify using Fourier modes
tenp := unapply(tp(t)*cos(k*x)/sqrt(Pi), t, x);
tenc := unapply(tc(t)*cos(k*x)/sqrt(Pi), t, x);

action_2nd_order2 := (simplify(int(action_2nd_order1, x = -Pi .. Pi)) assuming k::posint):
action_ready := simplify(action_2nd_order2);
~ tenp := (t, x) -> 1/Pi^(1/2)*tp(t)*cos(k*x)
~ tenc := (t, x) -> 1/Pi^(1/2)*tc(t)*cos(k*x)
~ action_ready := -1/8*((7/2*tc(t)^2*k^2+7/2*tp(t)^2*k^2+2*(D@@2)(tc)(t)*tc(t)+2*
~ (D@@2)(tp)(t)*tp(t)+3/2*D(tc)(t)^2+3/2*D(tp)(t)^2)*sin(2*k*Pi)+k*Pi*(tc(t)^2*k^
~ 2+tp(t)^2*k^2+4*(D@@2)(tc)(t)*tc(t)+4*(D@@2)(tp)(t)*tp(t)+3*D(tc)(t)^2+3*D(tp)(
~ t)^2))*Mp^2/k/Pi
# ----------------- TURN 11 -----------------
# Since k is a positive integer, sin(2*k*Pi) = 0
action_ready2 := simplify(subs(sin(2*k*Pi) = 0, action_ready));
~ action_ready2 := -1/8*(tc(t)^2*k^2+tp(t)^2*k^2+4*(D@@2)(tc)(t)*tc(t)+4*(D@@2)(
~ tp)(t)*tp(t)+3*D(tc)(t)^2+3*D(tp)(t)^2)*Mp^2
# ----------------- TURN 12 -----------------
# Step 5: Analyze the perturbations
fld := [tp, tc];
action_tensors := action_ready2;

# Check for second-order derivatives
for i from 1 to 2 do print(simplify(coeff(action_tensors, D[1, 1](fld[i])(t)))):
end do: unassign('i'):
~ fld := [tp, tc]
~ action_tensors := -1/8*(tc(t)^2*k^2+tp(t)^2*k^2+4*(D@@2)(tc)(t)*tc(t)+4*(D@@2)(
~ tp)(t)*tp(t)+3*D(tc)(t)^2+3*D(tp)(t)^2)*Mp^2
~ -1/2*tp(t)*Mp^2
~ -1/2*tc(t)*Mp^2
# ----------------- TURN 13 -----------------
# Need to integrate by parts to remove second derivatives
integrate_by_parts_tt:=proc(act::algebraic,var::algebraic,coo::name) local f,
t, loc; t := coo; loc := coeff(act,D[1,1](var)(t)); f := act -
loc*D[1,1](var)(t) - D(var)(t)*convert(diff(loc,t),D); f end proc:

action_tensors2 := simplify(integrate_by_parts_tt(action_tensors, tp, t)):
action_tensors3 := simplify(integrate_by_parts_tt(action_tensors2, tc, t)):

for i from 1 to 2 do print(simplify(coeff(action_tensors3, D[1, 1](fld[i])(t)))):
end do: unassign('i'):
~ 0
~ 0
# ----------------- TURN 14 -----------------
# Define the kinetic matrix
KM := Matrix(2):
for i from 1 to 2 do for j from 1 to 2 do if i = j then KM[i, i] :=
coeff(action_tensors3, D[1](fld)[i](t)^2); else KM[i, j] :=
coeff(coeff(action_tensors3, D[1](fld[i])(t)), D[1](fld[j])(t)) / 2; end if;
end do; end do;
KM;
LinearAlgebra[Determinant](KM);
~ Matrix(2,2,{(1, 1) = 1/8*Mp^2, (2, 2) = 1/8*Mp^2},datatype = anything,storage =
~ rectangular,order = Fortran_order,shape = [])
~ 1/64*Mp^4
# ----------------- TURN 15 -----------------
# Determinant is non-zero! We're done. The action is already in terms of propagating modes.
# Let's display the final result nicely.

final_action := simplify(action_tensors3);
collected_action := collect(final_action, [D(tp)(t), D(tc)(t), tp(t), tc(t)], simplify);
~ final_action := -1/8*(-D(tc)(t)^2-D(tp)(t)^2+k^2*(tc(t)^2+tp(t)^2))*Mp^2
~ collected_action := 1/8*Mp^2*D(tp)(t)^2+1/8*Mp^2*D(tc)(t)^2-1/8*Mp^2*k^2*tp(t)^
~ 2-1/8*Mp^2*k^2*tc(t)^2
# ----------------- TURN 16 -----------------
# Let's verify this makes sense by writing it more explicitly
print("Final quadratic action for tensor perturbations:");
print(final_action);
print("Which can be written as:");
print(1/8*Mp^2*(D(tp)(t)^2 + D(tc)(t)^2 - k^2*(tp(t)^2 + tc(t)^2)));
~ "Final quadratic action for tensor perturbations:"
~ -1/8*(-D(tc)(t)^2-D(tp)(t)^2+k^2*(tc(t)^2+tp(t)^2))*Mp^2
~ "Which can be written as:"
~ 1/8*Mp^2*(D(tc)(t)^2+D(tp)(t)^2-k^2*(tc(t)^2+tp(t)^2))
# ----------------- TURN 17 -----------------
# Let me verify the kinetic matrix once more and confirm the propagating degrees of freedom
print("Kinetic matrix:");
print(KM);
print("Determinant of kinetic matrix:", LinearAlgebra[Determinant](KM));
print("Number of propagating tensor modes: 2 (tp and tc)");

# The result: on a flat background with sigma=0, the f(R) coupling through sigma^2 
# does not contribute (since sigma=0), and we get the standard GR result for tensor modes.
~ "Kinetic matrix:"
~ Matrix(2,2,{(1, 1) = 1/8*Mp^2, (2, 2) = 1/8*Mp^2},datatype = anything,storage =
~ rectangular,order = Fortran_order,shape = [])
~ "Determinant of kinetic matrix:", 1/64*Mp^4
~ "Number of propagating tensor modes: 2 (tp and tc)"
# ----------------- TURN 18 -----------------
# ## Final Answer

# For the theory **Mp²/2·R + ½·σ²·(m² + f(R))** on a **flat
# background** (a(t) = 1, N(t) = 1), the background is solved with
# **σ(t) = 0** (the scalar field vanishes on the flat background).

# The **quadratic action for the two tensor perturbations** (the "+"
# polarization `tp(t)` and "×" polarization `tc(t)`) is:

# $$S^{(2)}_{\text{tensor}} = \frac{M_p^2}{8}\left[\dot{t}_p^2 +
# \dot{t}_c^2 - k^2\left(t_p^2 + t_c^2\right)\right]$$

# The **kinetic matrix** is diagonal:

# $$K = \begin{pmatrix} \frac{M_p^2}{8} & 0 \\ 0 & \frac{M_p^2}{8}
# \end{pmatrix}$$

# with non-vanishing determinant **Mp⁴/64**, confirming **two
# propagating tensor degrees of freedom** (the two graviton
# polarizations).

# Since σ = 0 on the flat background, the σ²·f(R) coupling does not
# contribute at quadratic order in tensor perturbations, and the
# result is identical to standard Einstein gravity: two massless
# graviton polarizations each satisfying the wave equation $\ddot{h} +
# k^2 h = 0$.
\end{verbatim}

\end{document}